\documentclass{article}

\usepackage[preprint]{corl_2022} %

\title{Rethinking Sim2Real: \emph{Lower} Fidelity Simulation Leads to \emph{Higher} Sim2Real Transfer in Navigation}

\usepackage{amssymb}
\usepackage{booktabs}
\usepackage{makecell}
\usepackage[ruled,noend,linesnumbered]{algorithm2e}
\usepackage{titlesec}
\usepackage{wrapfig}
\usepackage{graphicx}
\usepackage{array}
\usepackage{multirow}
\usepackage{adjustbox}
\usepackage{colortbl}
\usepackage{tabularx}
\usepackage{nicefrac}
\usepackage{macros/mysymbols}

\usepackage{grffile}
\usepackage{xspace}
\usepackage{needspace}

\usepackage{comment} 
\usepackage{url}
\usepackage{paralist}

\usepackage[belowskip=0pt,aboveskip=5pt,font=small]{caption}
\usepackage[belowskip=0pt,aboveskip=0pt,font=small]{subcaption}
\usepackage{xcolor}
\newcommand{\xhdr}[1]{\vspace{2pt}\noindent\textbf{#1}}

\newcommand{\pointnav}[1]{PointGoal Navigation\xspace}

\newcommand{\tableTitle}[1]{\normalsize{#1}}%
\newlength{\figwidth}%

\newcolumntype{Y}{>{\centering\arraybackslash}X}
\newcolumntype{P}[1]{\begin{center}>{\arraybackslash}p{#1}\end{center}}

\author{
  Joanne Truong$^{1}$, Max Rudolph$^{1}$, Naoki Yokoyama$^{1}$, \\
  \textbf{Sonia Chernova$^{1}$, Dhruv Batra$^{1,2}$, Akshara Rai$^{2}$}\\
  $^{1}$Georgia Institute of Technology, $^{2}$Meta AI  \\
  \texttt{\{truong.j, maxrudolph, nyokoyama, chernova, dbatra\}@gatech.edu} \\
  \texttt{\{akshararai\}@fb.com} \\
}

\begin{document}
\maketitle

\begin{abstract}
    If we want to train robots in simulation before deploying them in reality, it 
    seems natural and almost self-evident to presume that reducing the sim2real gap involves creating simulators of increasing fidelity (since reality is what it is). 
    We challenge this assumption and present a contrary hypothesis -- sim2real transfer of robots may be improved with \emph{lower} (not higher) fidelity simulation. We conduct a systematic large-scale evaluation of this hypothesis on the problem of visual navigation -- in the real world, and on 2 different simulators (Habitat and iGibson) using 3 different robots (A1, AlienGo, Spot). Our results show that, contrary to expectation, adding fidelity does not help with learning; performance is poor due to slow simulation speed (preventing large-scale learning) and overfitting to inaccuracies in simulation physics. Instead, building simple models of the robot motion using real-world data can improve learning and generalization. %

\end{abstract}

\keywords{Sim2Real, Deep Reinforcement Learning, Visual-Based Navigation} 

\section{Introduction}

The sim2real paradigm consists of training robots in simulation (potentially for billions of simulation steps corresponding to decades of experience \cite{ddppo}) before deploying them in reality. The last few years have seen significant investments -- the development of new simulators \cite{habitat19iccv, szot2021habitat, shen2021igibson, isaac, deitke2020robothor, gan2020threedworld, james2020rlbench, xiang2020sapien, todorov2012mujoco, freeman2021brax, coumans2016pybullet}, curation and annotation of 3D scans and assets \cite{ramakrishnan2021habitat, chang2017matterport3d, chang2015shapenet}, and development of techniques for overcoming the sim2real gap \cite{truong2021bi, chebotar2019closing, peng2018sim, tobin2017domain} -- resulting in a number of successful demonstrations of sim2real transfer \cite{habitatsim2real20ral, sct21iros, truong2020learning, locomotion_terrain, locomotion_wild, kumar2021rma}.
However, no simulator is a perfect replica of reality and the main challenge in this paradigm is overcoming the sim2real gap, defined as the drop in a robot's performance in the real-world (compared to simulation). It seems natural and almost self-evident to presume that reducing this sim2real gap involves creating simulators of increasing physics fidelity, and this sometimes forms the default operating hypothesis of the field. %

We challenge this convention and present a counter-intuitive idea -- sim2real transfer of robots may be improved not by increasing but by \emph{decreasing} simulation fidelity.
Specifically, we propose that instead of training robots entirely in simulation, we use classical ideas from hierarchical robot control \cite{garrett2020integrated} to decompose the policy into a `high-level policy' (that is trained solely in simulation) and a `low-level controller' (that is designed entirely on hardware and may even be a black-box controller shipped by a manufacturer). This decomposition means that the simulator does not need to model low-level dynamics, which can save both simulation time (since there is no need to simulate expensive low-level controllers), and developer time spent building and designing these controllers. 

We conduct a systematic large-scale evaluation of our hypothesis on the task of PointGoal (visual) Navigation \cite{anderson2018evaluation} in unknown environments -- using 2 simulators (Habitat and iGibson) and 3 different robots (A1, AlienGo, Spot). We train policies using two physics fidelities -- kinematic and dynamic. Kinematic simulation uses abstracted physics and `teleports’ the robot to the next state using Euler integration; kinematic policies command robot center-of-mass (CoM) linear and angular velocities. Dynamic simulation consists of rigid-body mechanics and simulates contact dynamics (via Bullet \cite{coumans2016pybullet}); dynamic policies command CoM linear and angular velocities, which are converted to robot joint-torques by a low-level controller operating at 240 Hz. We find that across all robots, a kinematically trained policy outperforms dynamic policies, \emph{even when evaluated using dynamic simulation and control}. Additionally, we show that the trained kinematic policy can be transferred to a real Spot robot, which ships with manufacturer-provided `black-box' low-level controllers that cannot be accurately simulated. 
In contrast, dynamic policies fail to achieve efficient navigation behavior on Spot, due to the sim2real gap and less simulation experience. 

\begin{figure*}[t]
	\begin{center}
		\adjustbox{trim={0} {0} {0} {0},clip}%
		{\includegraphics[width=0.95\linewidth]{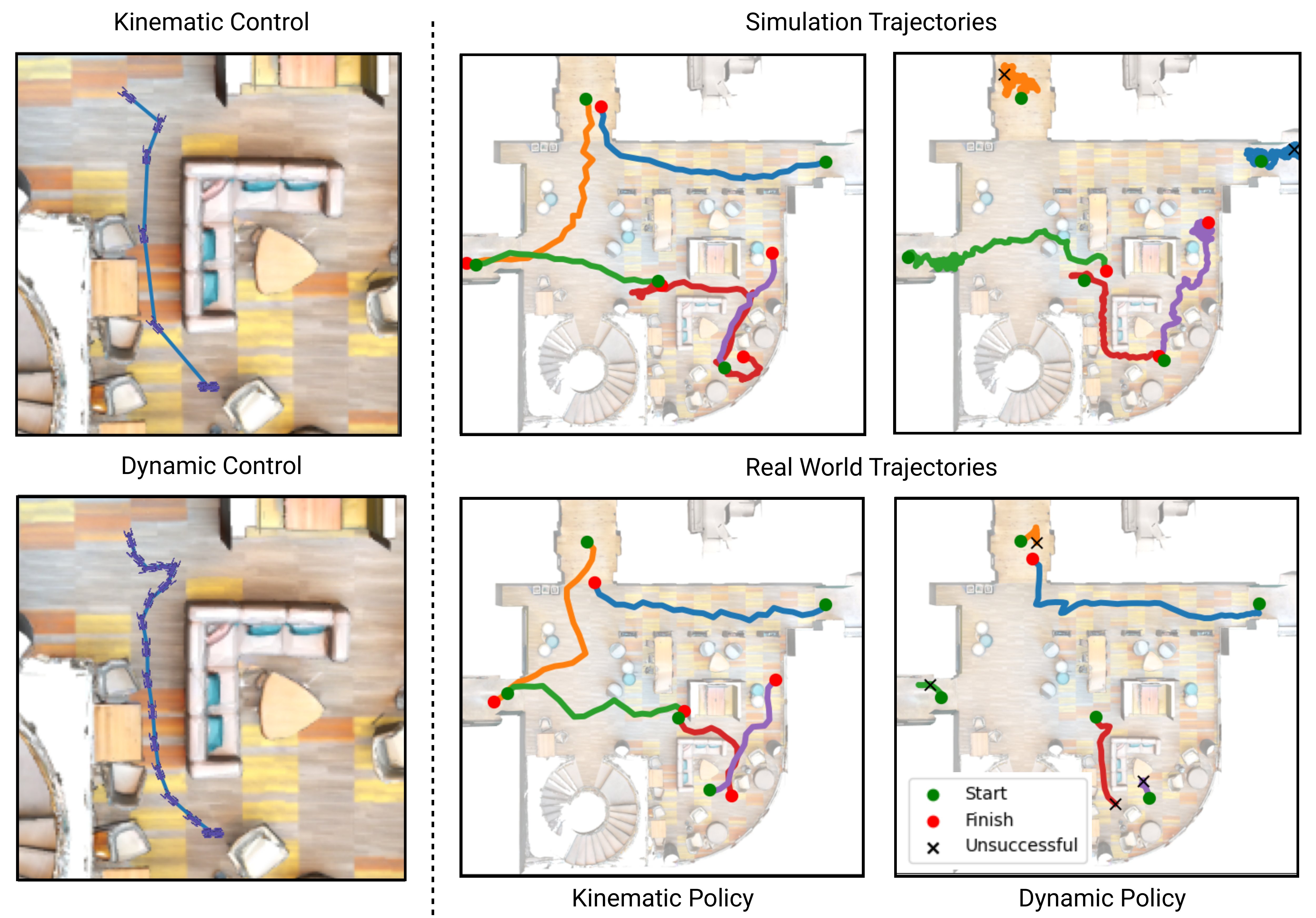}}
	\end{center}
	\looseness=-1
	\caption{\textbf{Left:} We train visual navigation policies at two levels of fidelity -- kinematic and dynamic. In kinematic control (top), the robot is `teleported' to the next state using Euler integration. In dynamic control (bottom), the robot's velocity commands are converted to leg joint-torques and rigid-body physics is simulated at 240Hz. \textbf{Right:} We evaluate the kinematic and dynamic trained policies in simulation (top) and the real-world (bottom) across 5 identical episodes. The kinematic policy achieves a $100\%$ success rate in all 5 episodes, and the robot takes similar paths in both simulation and the real-world. On the other hand, the dynamic policy achieves a 20-60$\%$ success rate, and the trajectories taken in simulation and the real-world do not correlate, pointing towards a larger sim2real gap. 
	}
	\label{fig:teaser}
\end{figure*}

The reasons for these improvements are perhaps unsurprising in hindsight -- learning-based methods overfit to simulators, and present-day physics simulators have approximations and imperfections that do not transfer to the real-world. A second equally significant mechanism is also in play -- lower fidelity simulation is typically faster, enabling policies to be trained with more experience under a fixed wall-clock budget. Even when the kinematic policies were trained for 2.3$\times$ less wall-clock time than the dynamic policies (with the same compute), the kinematic policies were able to learn from 10$\times$ the amount of data.
While our results are presented on legged locomotion and visual navigation, the underlying principle -- of architecting hierarchical policies and only training the high-level policy in an abstracted simulation -- is broadly applicable. We hope that our work leads to a rethink in how the research community pursues sim2real and in how we develop the simulators of tomorrow. Specifically, our findings suggest that instead of investing in higher-fidelity physics, the field should prioritize simulation speed for tasks that can be represented with abstract action spaces.

\vspace{-0.2cm}
\section{Related Work}
\vspace{-0.2cm}
\xhdr{Visual Navigation.} Recent works have shown that large-scale indoor environments and simulators like Habitat \cite{habitat19iccv, szot2021habitat} and iGibson \cite{shen2021igibson} can enable end-to-end learning of navigation policies from large amounts of agent- or expert-generated data \cite{ramrakhya2022habitat, chen2019learning, chaplot2020learning} on simple, wheeled systems. This is in contrast to the typical mapping and planning paradigm used in classical robotics, which can suffer when the quality of maps is low \cite{bansal2020combining} or requires expensive equipment like LiDAR \cite{hess2016real}. %
In this work, we show that such end-to-end learning is also possible for complex, legged robots. %

\xhdr{Sim2real for Legged Robots.} Sim2real quadrupedal locomotion has been widely studied in the past several decades \cite{truong2020learning, brain, mso, us, li2021planning}, with most learning low-level skills in simulation and transferring them to hardware \cite{tan2018sim}, or adapting them online to reduce the sim2real gap \cite{peng2020learning, rai2018bayesian}. However, these policies are typically blind, and use only proprioceptive sensors on the robot to determine actions \cite{locomotion_terrain, kumar2021rma}. In contrast, an autonomous robot needs to respond to its environment, and take visual input into account. Some works have proposed learning visual policies in simulation and applying them to the real-world \cite{truong2020learning, locomotion_wild, fu2021coupling}, and other works leverage expensive LiDAR sensors for external sensing \cite{rudin2022learning}. These works use learned or hand-designed physically simulated low-level controllers; we show that physics simulations can be detrimental to learning high-performing sim2real policies, even for complex legged robots. Work from \cite{hoeller2021learning} also utilize similar simulator simplifications to increase simulation speed in training navigation policies for 1 robot in a single room, but does not discuss how this formulation affects sim2real transfer. In our work we present a rigorous (multi-robot, multi-simulator) study of the effect of different simulation fidelities on visual navigation.

\xhdr{Abstracted Task-space Learning.} Abstracted (hierarchical, high-level) action spaces are common in robotics literature. Examples include task and motion planning for manipulation \cite{garrett2021integrated, kaelbling2011hierarchical, lin2022efficient}, legged locomotion \cite{drc, li2019using}, navigation \cite{habitatsim2real20ral}, etc. Several works reason over symbolic actions like pick and place, or hierarchical policies with discrete/continuous attributes \cite{zeng2020transporter, yuan2021sornet, brain, mso, us}, or even abstracted dynamics models \cite{li2021planning}. 
While the ideas of abstracted/hierarchical policies are fairly common, typically both the high- and low-level policies are learned in simulation and transferred to reality \cite{brain, mso, li2021planning}, often augmented with techniques like domain randomization \cite{tan2018sim} and real-word adaption \cite{peng2018sim}. 
Instead, we use an abstracted simulator, which does not model low-level physics, and learn high-level policies that are transferred to the real-world in a zero-shot manner. %

\vspace{-0.3cm}
\section{Experimental Setup}
\vspace{-0.2cm}
\label{sec:setup}
\xhdr{Task: PointGoal Navigation.} In the task of PointGoal Navigation \cite{anderson2018evaluation}, a robot is initialized in an unknown environment and is tasked with navigating to a goal coordinate without access to a pre-built map of the environment. The goal is specified relative to the robot's starting location for the episode (i.e., ``go to $\Delta$x, $\Delta$y"). The robot has access to an egocentric depth sensor and an egomotion sensor (sometimes referred to as GPS+Compass in this literature) from which the robot derives the goal location relative to its current pose. An episode is considered successful when the robot reaches the goal position within a success radius (typically half of the robot's body length). The robot operates within constraints of maximum number of steps per episode (150 for Spot) and velocity limits ($\pm$ 0.5 m/s for linear and $\pm$ 0.3 rad/s for angular velocities on Spot). We linearly scale the linear and angular velocity limits for A1 and Aliengo to be proportional to the length of each robot's leg, and inversely scale the maximum number of steps allowed. In effect, smaller robots have a smaller maximum allowed velocity to improve stability during execution, but are allowed more steps to reach the goal. The exact parameters used for each robot is described in the appendix. For evaluation, we report the success rate (SR), and Success inversely weighted by Path Length (SPL) \cite{anderson2018evaluation}, which measures the efficiency of the trajectory taken with respect to the ground-truth shortest path.

\xhdr{Robot Platforms.} We study visual navigation for 3 quadrupedal robots -- A1 and Aliengo from Unitree \cite{Unitree}, and Spot from Boston Dynamics (BD) \cite{spot} in simulation. In the real-world, we show sim2real transfer of the learned navigation policies to Spot. %
To have a consistent camera setup across all the robots, we attach an Intel RealSense D435 camera to Spot in the real-world, and use this camera for visual inputs to the policy. In our hardware experiments, we want to measure how often our sim2real policies lead to collisions without jeopardizing safety. We achieve this balance as follows: the BD collision-avoidance capability is kept turned on, set to trigger at a tight threshold of 0.10m. Next, we track the number of times the robot comes within 0.20m of any obstacle (as measured by any of the 5 onboard depth cameras). This gap (between 0.20m and 0.10m) allows us to record possible collisions while preventing actual ones. While the BD API allows for high-level navigation without access to a map, it cannot navigate around obstacles autonomously, without a map. In our work, we consider complex, long-range navigation paths (up to 30m) in cluttered environments with many obstacles; the goals are unreachable with the just BD navigation API.

\begin{figure*}[ht]
  \centering%
  \resizebox{\textwidth}{!}{
  \renewcommand{\tableTitle}[1]{\huge{#1}}%
  \setlength{\figwidth}{0.3\textwidth}%
  \setlength{\tabcolsep}{1.5pt}%
  \renewcommand{\arraystretch}{0.8}%
  \renewcommand\cellset{\renewcommand\arraystretch{0.8}%
  \setlength\extrarowheight{0pt}}%

  \hspace{-0.25cm}\begin{tabular}{c c c c}
   \tableTitle{\LARGE A1} & \tableTitle{\LARGE Aliengo} & \tableTitle{ \LARGE Spot} & \tableTitle{ \LARGE Spot-Real} \\
   \makecell{\includegraphics[height=0.17\textheight]{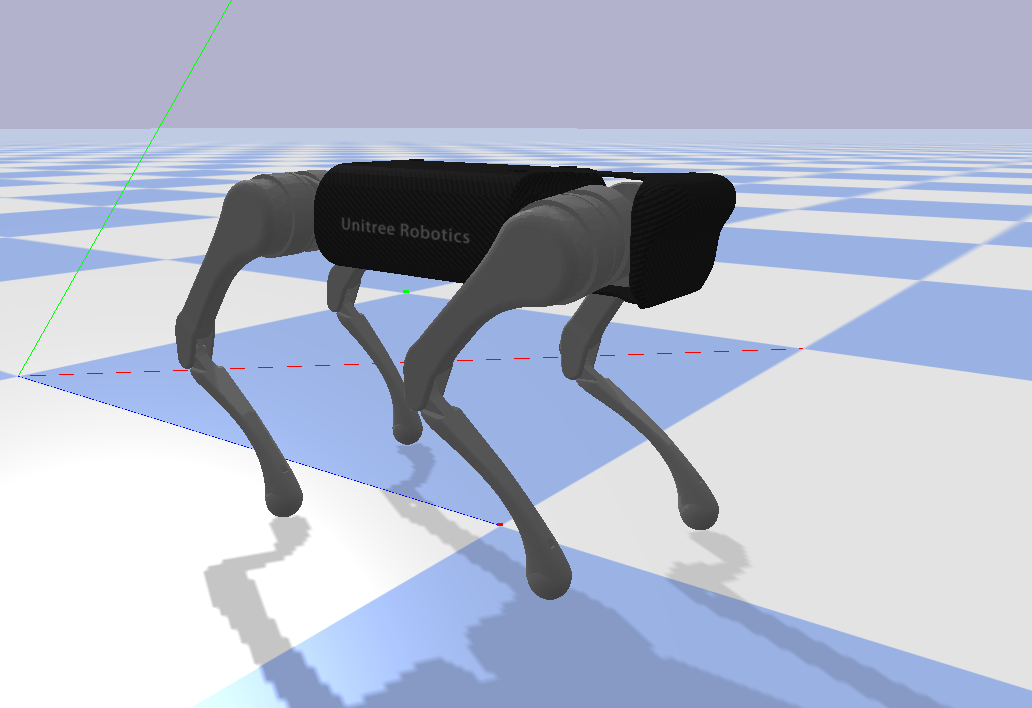}} &
   \makecell{\includegraphics[height=0.17\textheight]{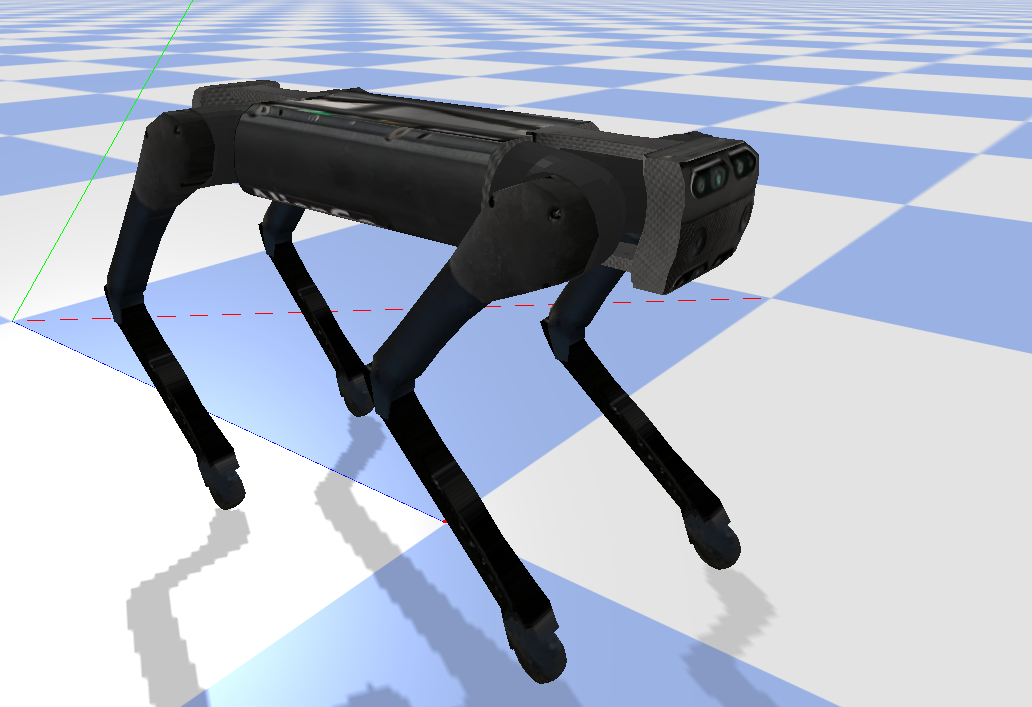}} & 
   \makecell{\includegraphics[height=0.17\textheight]{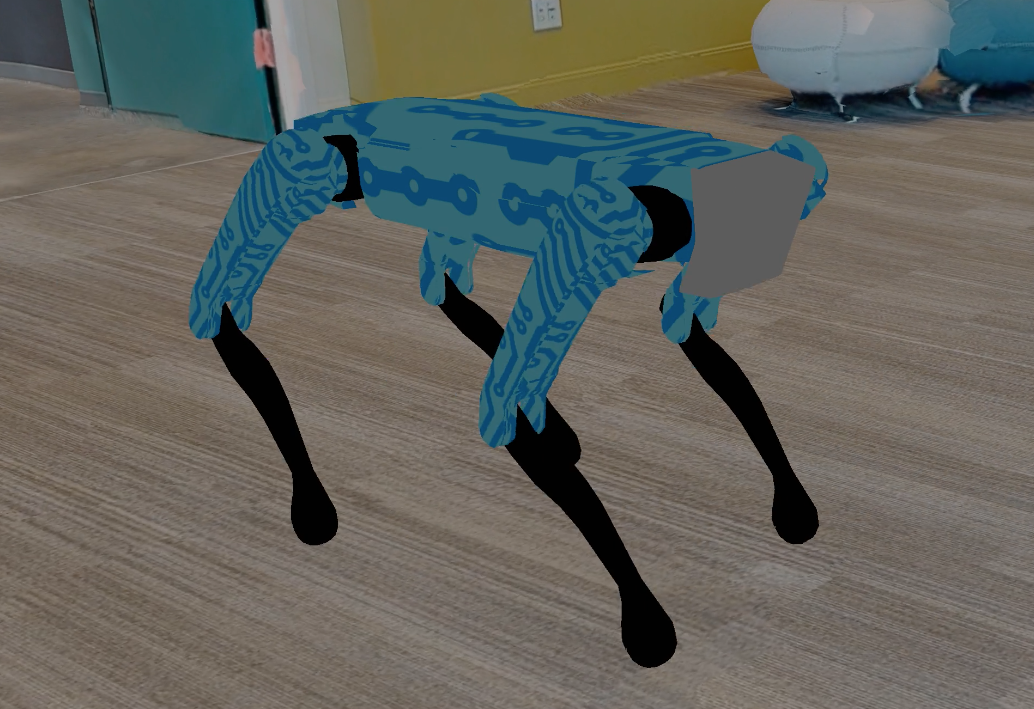}} &
   \makecell{\includegraphics[height=0.17\textheight]{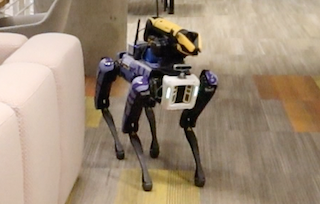}}
  \end{tabular}}
  \caption{\small Robots used for training and evaluation. 
  }
  \label{fig:robots}
\end{figure*}

\xhdr{Simulation Environments.} We use two simulation platforms -- Habitat \cite{habitat19iccv, szot2021habitat} and iGibson \cite{shen2021igibson} for training and evaluation. Both simulators support rendering of photorealistic environments; Habitat uses a low-level (C++) integration with the Bullet physics engine \cite{coumans2016pybullet}, while iGibson leverages PyBullet, the Python-based integration of Bullet. Thus, while the underlying physics engines between the two are the same, Habitat runs $\sim$ 1200\% faster than iGibson \cite{szot2021habitat}. This allows us to train policies faster with Habitat than with iGibson even when using identical policies and compute. %

\xhdr{Dataset.} For training and evaluation, we use a combination of the Habitat-Matterport (HM3D) \cite{ramakrishnan2021habitat} and Gibson \cite{xia2018gibson} 3D datasets. The two datasets combined consist of over 1000 high-resolution 3D scans of real-world indoors environments, and consists of realistic clutter. %
We generate training and evaluation episodes compatible with our robots for the HM3D and Gibson scenes following the procedure described in \cite{habitat19iccv}. Specifically, we restrict the geodesic distance from the start and positions to be between 1 and 30m, and increase navigation complexity by rejecting paths that consist of near-straight lines, with few obstacles. As described in \cite{habitat19iccv}, both of these heuristics result in complex, but navigable paths. Additionally, we check for collisions along the sampled paths using the URDF of the largest robot (Spot) to ensure that all paths are navigable.

\begin{wrapfigure}{r}{0.5\textwidth}
\vspace{-0.9cm}
  \begin{center}
    \includegraphics[width=0.4\textwidth]{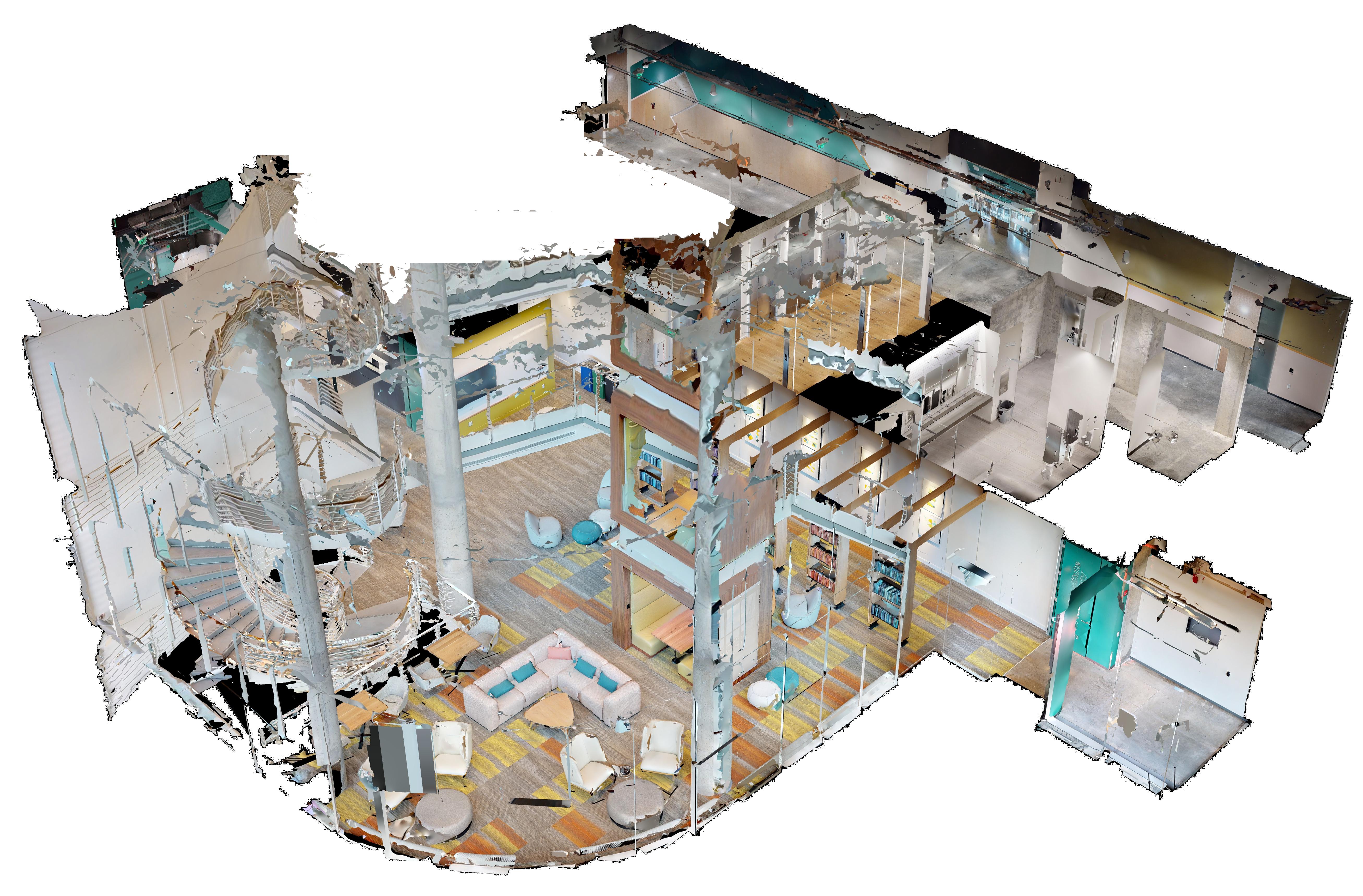}
    \caption{The real-world testing environment is a part of a large commercial building and contains clutter from furniture such as tables, bookshelves, and couches.}
    \vspace{-0.8cm}
    \label{fig:coda}
   \end{center}
\end{wrapfigure}

\looseness=-1
\xhdr{Real-World Test Environment.} The real-world evaluation environment, LAB, is a 325m$^2$ lobby in a commercial office building. The lobby contains furniture such a couches, cushions, bookshelves and tables. We specify a set of 5 waypoints as the start and end locations for the navigation episodes in LAB with an average episode length of 10m. %
We match the furniture layout to the position captured in the 3D scan (Figure \ref{fig:coda}) to run identical evaluation experiments in both simulation and the real-world. The scan of LAB is not part of training.

\vspace{-0.35cm}
\section{Kinematic and Dynamic Control for Visual Navigation}
\begin{figure*}[t]
	\begin{center}
		\adjustbox{trim={0} {0} {0} {0},clip}%
		{\includegraphics[width=0.9\linewidth]{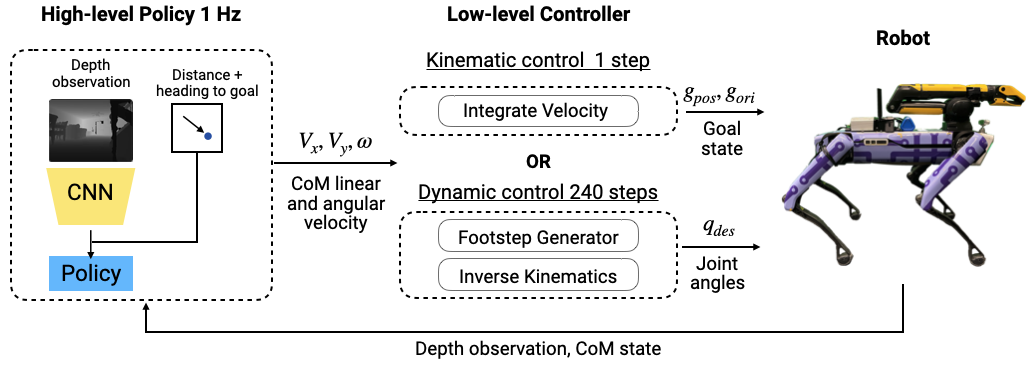}}
	\end{center}
	\caption{Our architecture for PointGoal Navigation on a legged robot. A high-level visual navigation policy predicts CoM linear and angular velocities. The velocities are passed into either a kinematic or dynamic low-level controller to step the robot in simulation. In the real-world, we directly send the velocity commands from the high-level policy to the robot, and uses the low-level controller from Boston Dynamics for movement.}
	\label{fig:architecture}
\end{figure*}

\vspace{-0.3cm}
\looseness=-1
As illustrated in Figure \ref{fig:architecture}, our proposed approach is hierarchical, with (1) a high-level visual navigation policy that commands desired center of mass (CoM) motion at 1Hz, and (2) a low-level controller that follows this desired motion. 
We consider controllers at two levels of abstraction -- `kinematic' and `dynamic'. The kinematic controller simply integrates the desired velocity and outputs a CoM position at 1Hz; kinematic simulation then teleports the robot to the desired state. The dynamic controller uses a low-level controller that commands joint torques at 240Hz; dynamic simulation models rigid-body and contact dynamics via Bullet (with a physics step-size of \nicefrac{1}{240} sec). We provide details of all three of these pieces (high-level policy, kinematic and dynamic controllers) next. 

\looseness=-1
\xhdr{High-level Visual Navigation Policies.}
The high-level policy takes as input an egocentric depth image, and the goal location relative to the robot's current pose. The output of the policy is a 3-dimensional vector, representing the desired CoM forward, lateral, and angular velocities ($V_x, V_y, \omega$). The neural network architecture consists of a ResNet-18 visual encoder and a 2-layer LSTM policy. Using a recurrent policy allows the policy to learn temporal dependencies through the hidden state. The final layer of the policy parameterizes a Gaussian action distribution from which the action is sampled. The policy is trained using DD-PPO \cite{ddppo}, a distributed reinforcement learning method, in both the Habitat and iGibson simulators. Our reward function is derived from \cite{truong2020learning}, with an added penalty for backward velocities, which can lead to collisions and hurts performance. 

\xhdr{Kinematic Control and Simulation.}
In kinematic control, the final state of the robot is calculated by integrating the desired CoM velocity commanded by the high-level navigation policy at 1Hz. The robot is directly moved to the desired pose, without running a physics simulation.
In both Habitat and iGibson, the robot is kept in place if being at the new desired state would result in a collision.

The objective of the kinematic control is to abstract away the low-level physics interactions between the robot and its environment. This has two advantages: (1) it avoids the need to accurately model low-level controllers, especially for closed-source robots like Spot; (2) it enables faster simulation speed by avoiding high-frequency physics integration, conducive to model-free RL that requires large amounts of experience. On the other hand, teleporting the robot to the desired state might remove necessary dynamics, such as poor tracking of low-level controllers. In Section \ref{sec:noise}, we propose how to incorporate such low-level characteristics into a kinematic simulation using real-world data.

\xhdr{Dynamic Control in Simulation and Hardware.} 
\looseness=-1
We experiment with two different low-level dynamic controllers for quadruped robots. The first is an expert-designed Raibert-style controller from \cite{truong2020learning}, which consists of a footstep generator and an inverse kinematic solver that commands desired joint angles from CoM velocities. The joint angles are converted to joint torques using a linear feedback controller, and applied to the simulation. This controller was shown to achieve sim2real transfer for A1 \cite{truong2020learning}. However, on other robots in our experiments, it shows relatively poor tracking of high-level commands. Thus, we also experiment with another model-predictive control (MPC) dynamic controller from \cite{peng2020learning}, which commands joint torques directly. This controller has been applied to real-world A1 robot \cite{yang2022fast, li2021model} and shows better tracking of desired velocities for our test robots, as compared to the Raibert controller from \cite{truong2020learning}. However, MPC is prohibitively slow and cannot be used for training RL policies. Thus, we use Raibert for training dynamic policies, but evaluate using MPC. \footnote{Evaluation using Raibert \cite{truong2020learning} can be found in the appendix.} This difference in train and evaluation dynamics controllers has multiple purposes: (1) the evaluation using MPC improves performance of most policies, including dynamic policies, due to its better ability to track high-level commands; (2) the difference between the two dynamic controllers in simulation is also a proxy for the difference between our low-level controllers and closed-source controllers from Spot. If a dynamic policy cannot transfer from Raibert to MPC, it has a low chance of transfer to Spot which has black box BD controllers, or even other robots in the real-world.

Both dynamic controllers model the low-level physics interactions between the robot and the environment. %
This makes them considerably slower than the kinematic controller, making training RL policies challenging. Moreover, for Spot, the low-level controller implementation is not openly available, making it hard to reproduce the low-level controller in simulation. For commercial legged robots that come with black-box controllers, kinematic simulations are the ideal fidelity for learning navigation policies. Our experiments in Section \ref{sec:results} show that the added fidelity of dynamic controllers does not benefit policy learning, or sim2real transfer.

\xhdr{Impact of Low-level Controllers on Policy Learning.}
The low-level controller used in dynamic simulations can have a significant impact on the learned policy. 
Low-level controllers both in sim and real are biased, and the status quo is to make them biased in the same way, as the policy learns to compensate for the bias error. For example, \cite{hwangbo2019learning} add learned actuation noise to their simulation, while \cite{tan2018sim} measure hardware characteristics and add them to the simulation. However, given the high-dimensional nature of low-level physics, it is very difficult to ensure that the biases incorporated in simulation actually hold for a larger range of motions that the data was collected on.
Thus, there are several iterations of data-collection, bias updates, training and deployment needed for good performance. Instead, kinematic controllers are unbiased by design, and can easily incorporate hardware bias through low-dimensional CoM motion noise models created from a small amount of real-world data, as shown in our experiments in Section \ref{sec:results}.

\section{Results and Analysis}
\label{sec:results}
In this section, we first study generalization of visual navigation policies across simulators (trained in one sim, tested in another) and across controllers (trained with one controller, tested with another). %
This shows the importance of fast simulation for learning high-level policies by comparing performance of kinematic and dynamic policies trained for the same wall-clock time. Next, we examine the performance of the different policies at zero-shot sim2real transfer on the Spot robot. %

\begin{figure}[h]
\vspace{-0.3cm}
  \centering%
  \resizebox{\columnwidth}{!}{
  \renewcommand{\tableTitle}[1]{\large{#1}}%
  \setlength{\figwidth}{0.27\columnwidth}%
  \setlength{\tabcolsep}{1.5pt}%
  \renewcommand{\arraystretch}{0.8}%
  \renewcommand{\cellset}{\renewcommand\arraystretch{0.8}%
  \setlength\extrarowheight{0pt}}%

  \hspace{-0.25cm}\begin{tabular}{c c c c}
   \makecell{\includegraphics[width=0.27\textwidth]{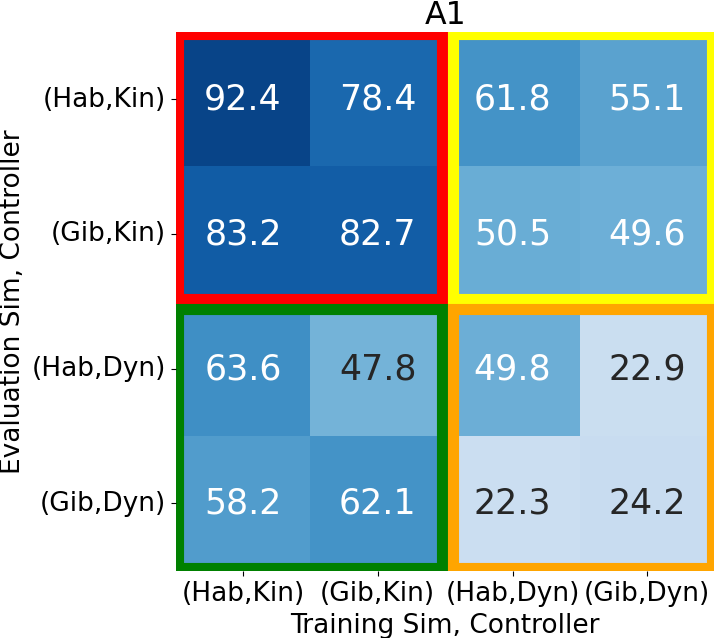}} &
   \makecell{\includegraphics[width=0.27\textwidth]{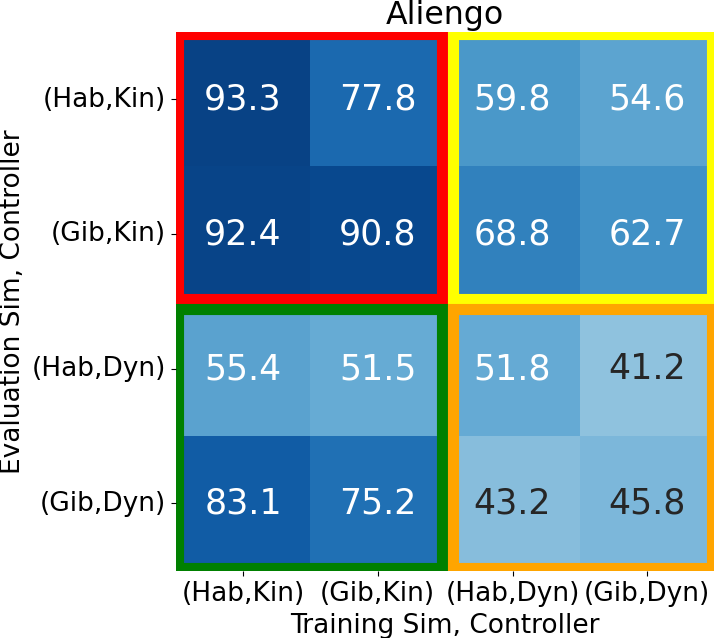}} &
   \makecell{\includegraphics[width=0.27\textwidth]{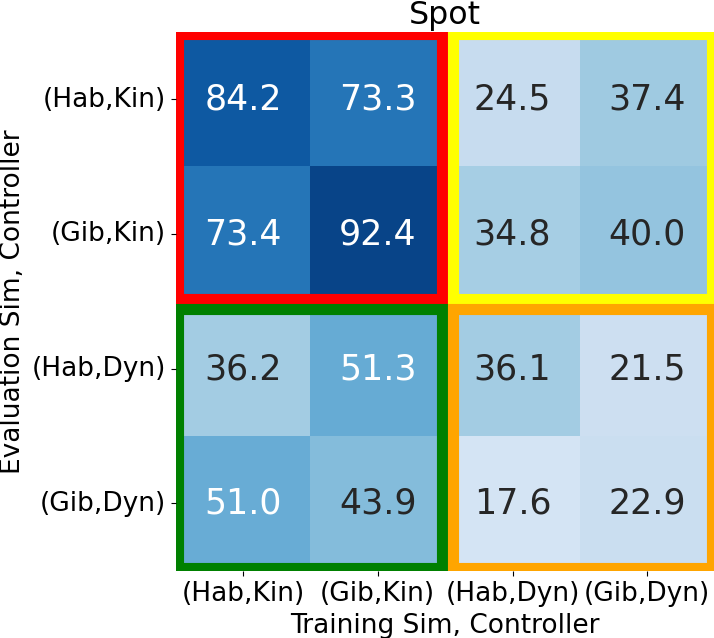}} &
   \makecell{\includegraphics[width=0.042\textwidth]{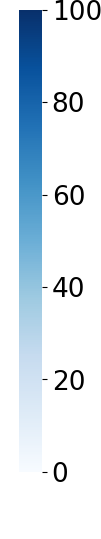}}
  \end{tabular}}
  \caption{Average success rates for sim2sim and kinematic2dynamic transfer for A1, Aliengo and Spot. We see that the kinematic trained policies perform the best overall (\textcolor{red}{red} quadrants), and also often outperform the dynamic trained policies, even when evaluated using dynamic control (\textcolor{green}{green} quadrants vs. \textcolor{orange}{orange} quadrants). 
} 
  \label{fig:heatmaps}
  \vspace{-0.5cm}
  \end{figure}

\xhdr{How large is the sim2sim gap? High for dynamic, and low for kinematic policies.}
\looseness=-1
We exhaustively study the combinatorial space of experiments -- policies trained under 2 training conditions (with kinematic and dynamic simulation) $\times$ 2 evaluation conditions (kinematic and dynamic simulation) $\times$ 2 simulators (Habitat and iGibson) $\times$ 3 robots (A1, Aliengo, Spot). 
For each condition, we train and report results with 3 random seeds. 
Each policy is trained using 8 GPUs for 3 days, 
resulting in a cumulative training budget of 6,912 GPU-hours (288 GPU-days). 
The average success rates are presented in Figure \ref{fig:heatmaps}. Rows represent the evaluation conditions as tuples (simulator, fidelity), while columns represent the training conditions. We evaluate all policies across 1,100 episodes from 110 unique scenes in the HM3D + Gibson validation split. 

We make two key observations here: 
\begin{asparaenum}
   \item \textbf{Kinematic-trained policies perform best overall, for all robots.} In all cases, kinematic policies outperform the dynamic policies, \emph{even when evaluated using dynamic control}, \eg \textcolor{green}{62.1\% SR} for A1 in (iGibson, Kinematic) \vs \textcolor{orange}{24.2\% SR} in (iGibson, Dynamic), Fig. \ref{fig:heatmaps}, left. This is a surprising result because the kinematic policies are being evaluated in an out-of-distribution setting, which was never seen or accounted for during training. On the other hand, the dynamic policies are being evaluated in the domain that they were trained in, hence do not require control-related generalization.
   \item \textbf{Dynamic policies are not robust to different dynamic simulations.} The dynamic policies from the two simulations observe significant performance drops when evaluated in the other dynamic simulation. This points to the dynamic policies overfitting to the simulator dynamics during training, failing to generalize to a new setting, see \eg column 3, rows 3 and 4; \textcolor{orange}{49.8\% SR} for A1 in (Habitat, Dynamic) vs. \textcolor{orange}{22.3\% SR} in (iGibson, Dynamic). (iGibson, Dynamic) shows poor performance in both iGibson and Habitat, with slightly poorer performance in Habitat. This sensitivity to simulation makes training dynamics policies difficult, especially when the controller for the real-world robot is unknown. Even if the real-world controller is known, simulation physics and real-world are different, and sim2real transfer of the learned policy can suffer (as evidenced by low sim2sim transfer). On the other hand, kinematic policies, that have been trained with no physics, can generalize to the different dynamic controllers. Both of these results go to show that not only kinematic trained policies are able to learn the task well, they have learned to reason without overfitting to simulation physics, making their chances of successful sim2real transfer high. 
 \end{asparaenum}

\begin{figure}[h!]
  \vspace{-0.2cm}
  \centering%
  \resizebox{\columnwidth}{!}{
  \renewcommand{\tableTitle}[1]{\large{#1}}%
  \setlength{\figwidth}{0.3\columnwidth}%
  \setlength{\tabcolsep}{1.5pt}%
  \renewcommand{\arraystretch}{0.8}%
  \renewcommand{\cellset}{\renewcommand\arraystretch{0.8}%
  \setlength\extrarowheight{0pt}}%

  \hspace{-0.25cm}\begin{tabular}{c c}
   \makecell{\includegraphics[width=0.44\textwidth]{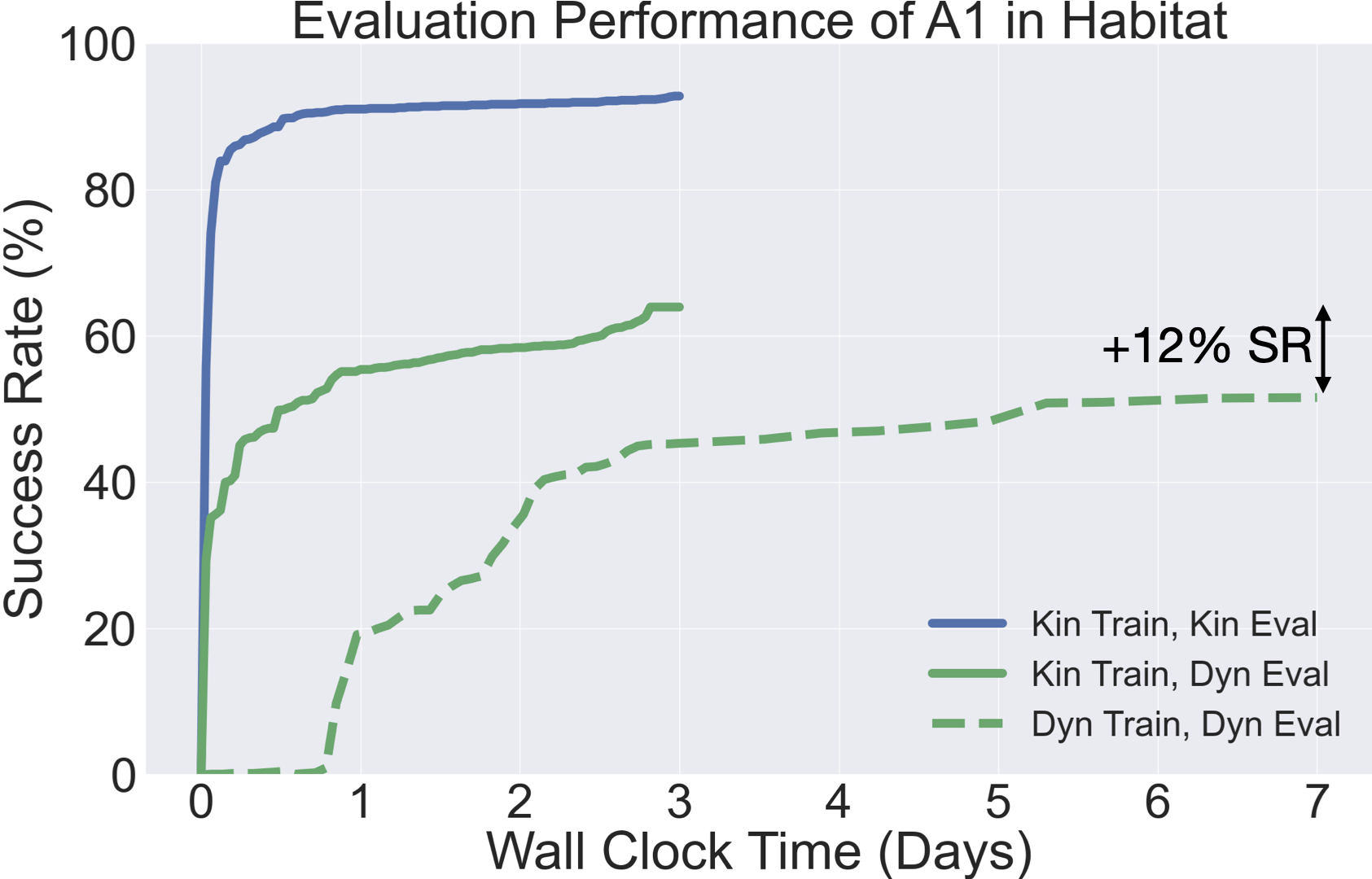}} &
   \makecell{\includegraphics[width=0.44\textwidth]{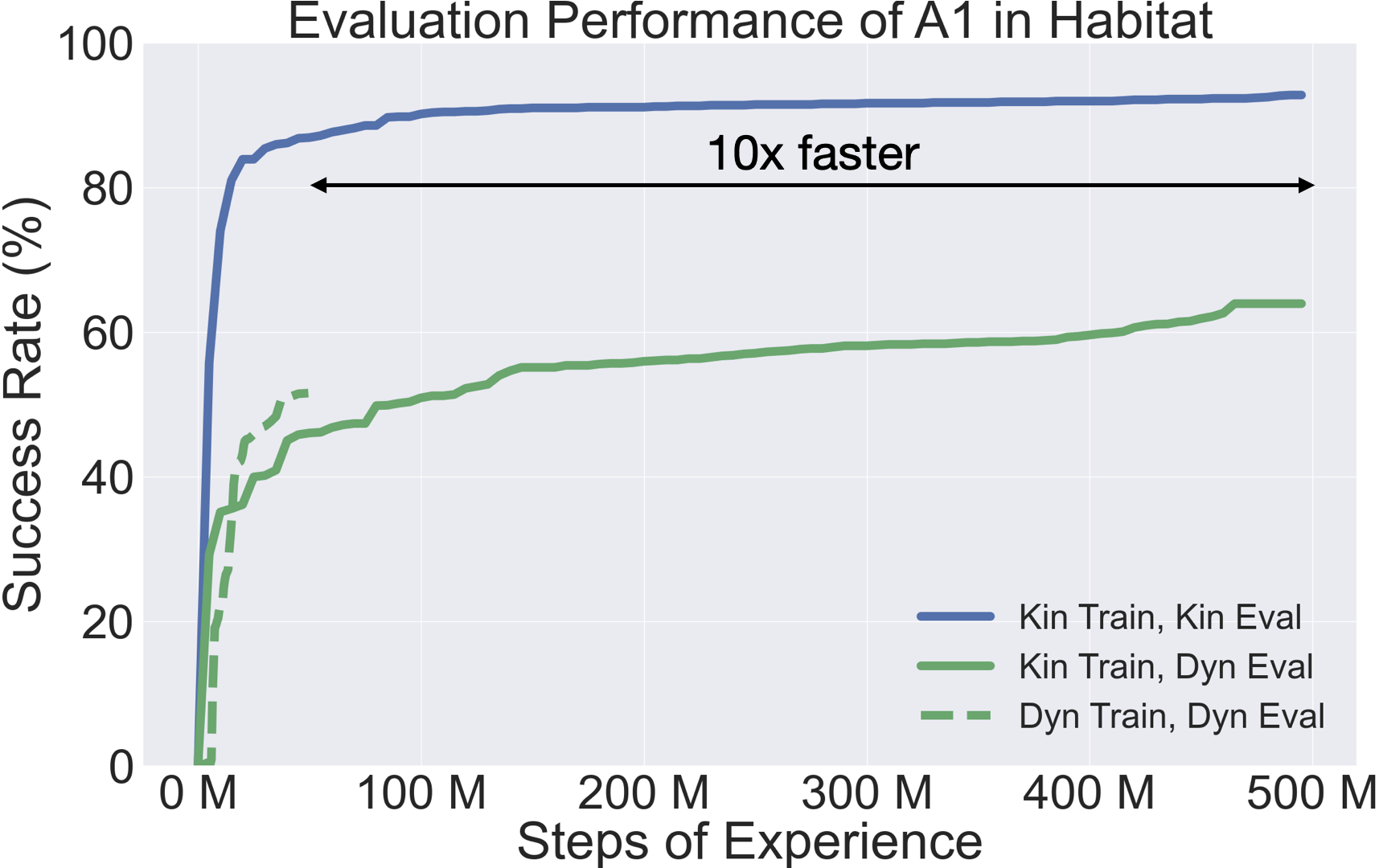}} 
  \end{tabular}}
  \caption{Success rate of PointNav policies with A1 trained and evaluated in Habitat with kinematic or dynamic control. Left: Kinematic
policies outperform the dynamic trained policies (+12\% SR), even when evaluated using dynamic control. Right: Using kinematic control, we can train our robot for 10$\times$ more steps of experience than with dynamic control under identical compute budgets, despite training for 2.3$\times$ less wall-clock time.}
  \label{fig:plot}
  \vspace{-0.5cm}
  \end{figure}

\xhdr{Why do kinematic-trained policies outperform dynamic ones? Scale.} 
We plot the evaluation performance of both policies in Habitat kinematic and dynamic simulation in Figure \ref{fig:plot}. We train both policies to convergence-- 3 days for the kinematic policies, and 7 days for the dynamic policies. While the kinematic policies are trained for 2.3$\times$ less wall-clock time, they still outperform the longer trained dynamic policies, even when evaluated out-of-distribution using dynamic control (+12\% SR). Kinematic training is much faster than training dynamically (right, Fig. \ref{fig:plot}); with kinematic training, the robot is able to learn from approximately 10$\times$ more steps of experience (500M steps vs. 50M steps). This increased experience allows the kinematic policies to learn intelligent high-level reasoning. We contend that for any computational budget, there will always be more complex tasks that are bottlenecked by that budget. Wall-clock time is the true limitation for learning-based sim2real approaches (not experience, as different simulators have different speeds).

\begin{figure}[t]
    \centering
    \includegraphics[width=0.31\textwidth]{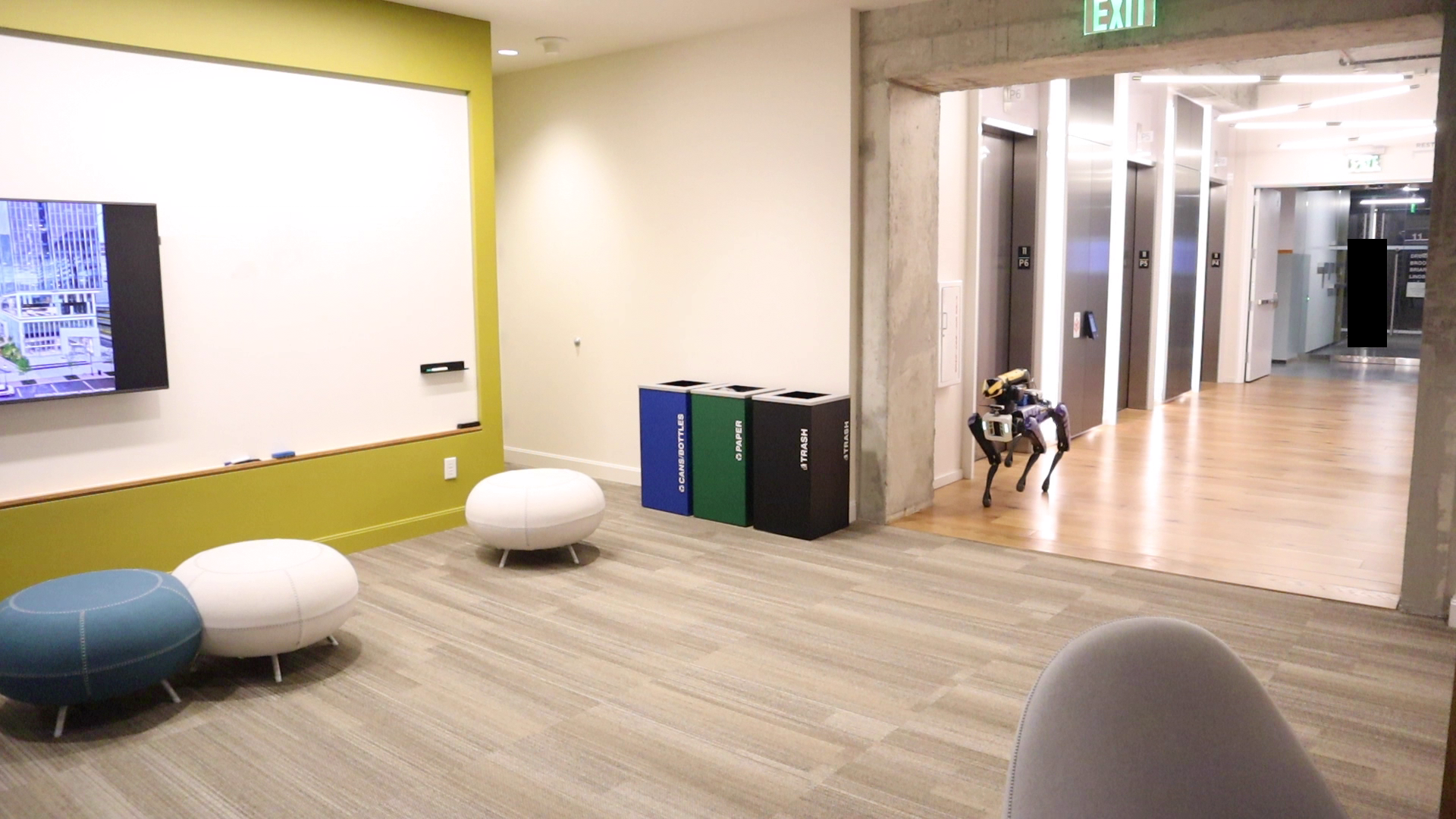}
    \includegraphics[width=0.31\textwidth]{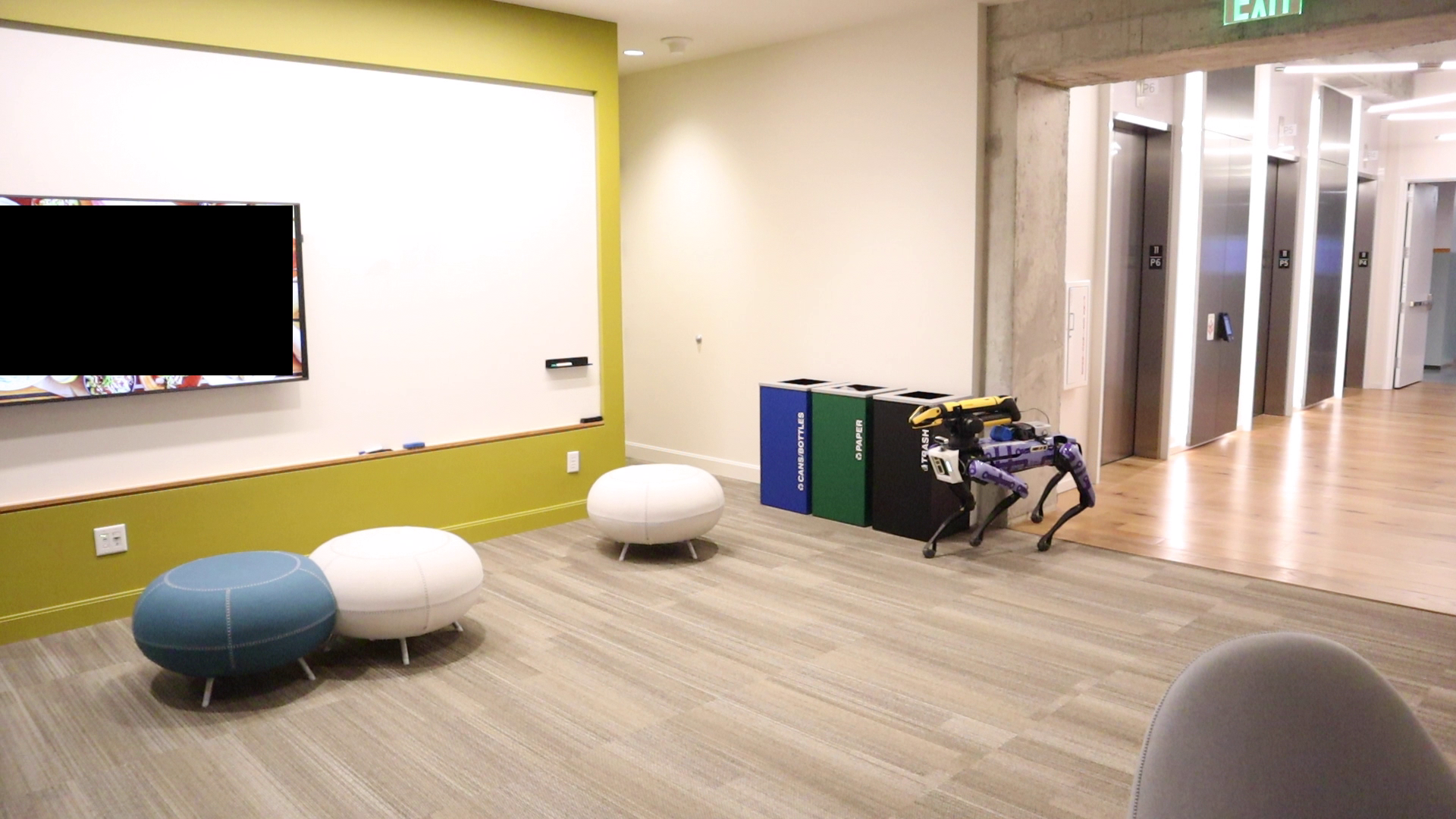}
    \includegraphics[width=0.31\textwidth]{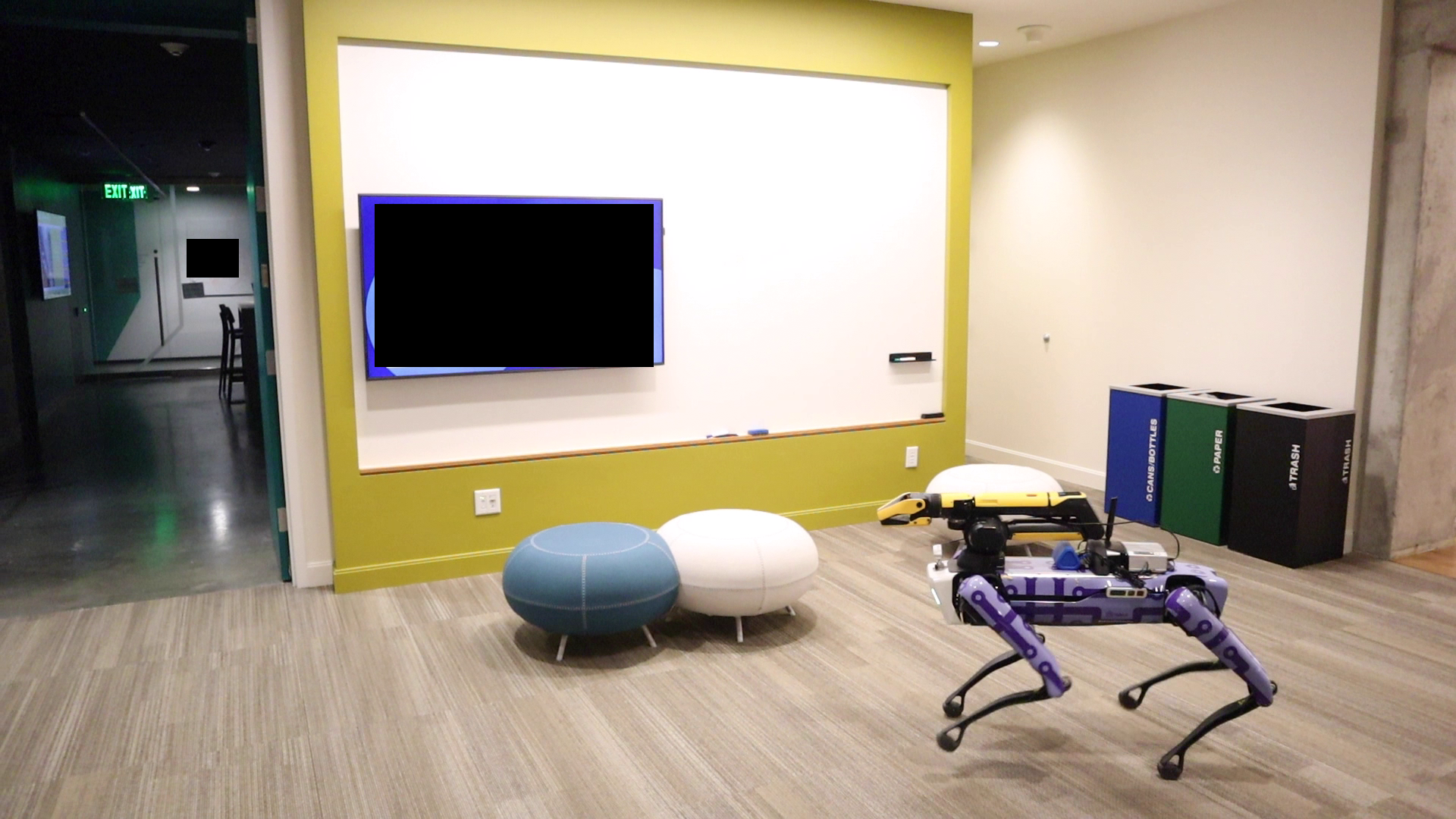}
    \\
    \includegraphics[width=0.31\textwidth]{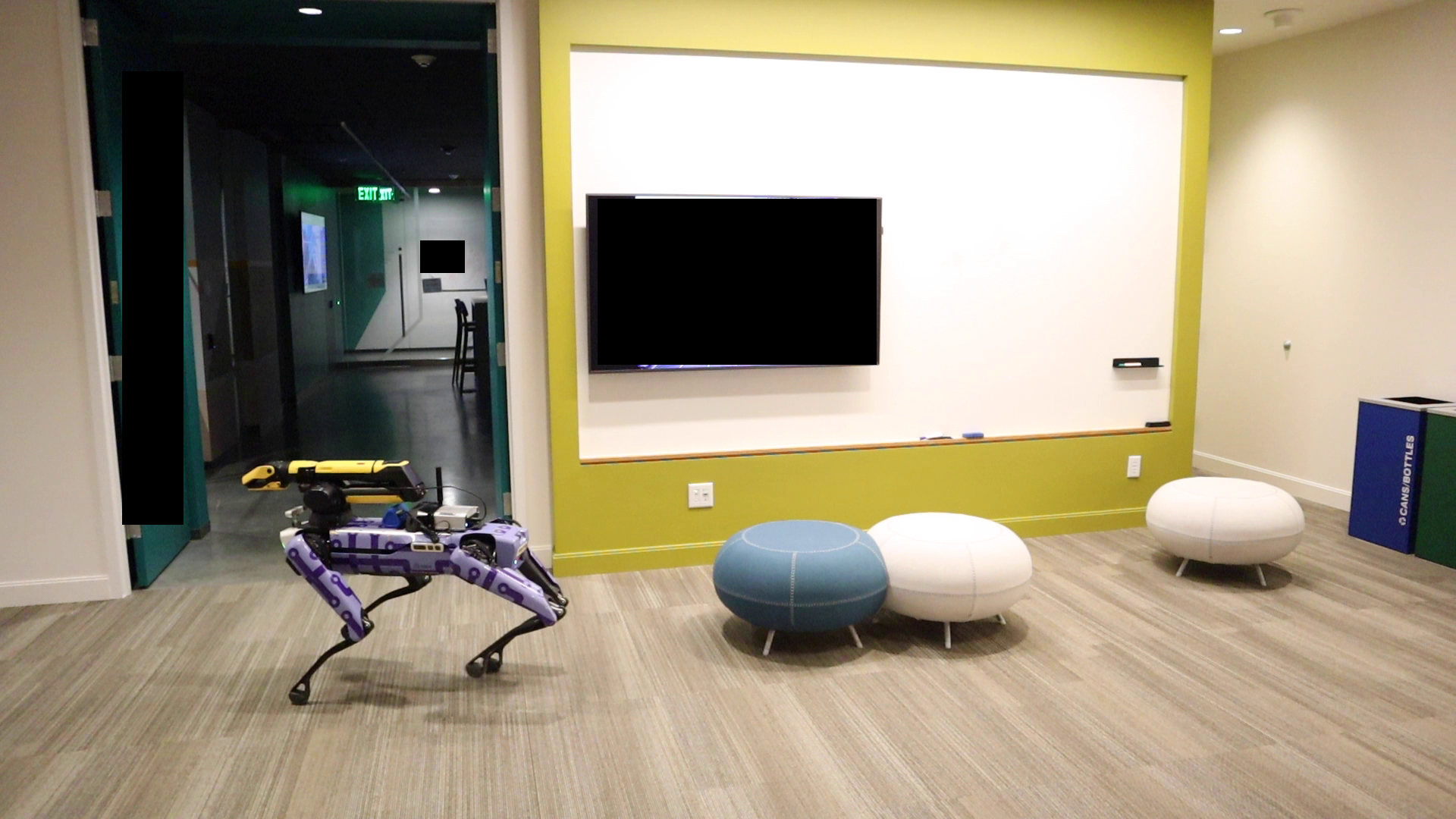}
    \includegraphics[width=0.31\textwidth]{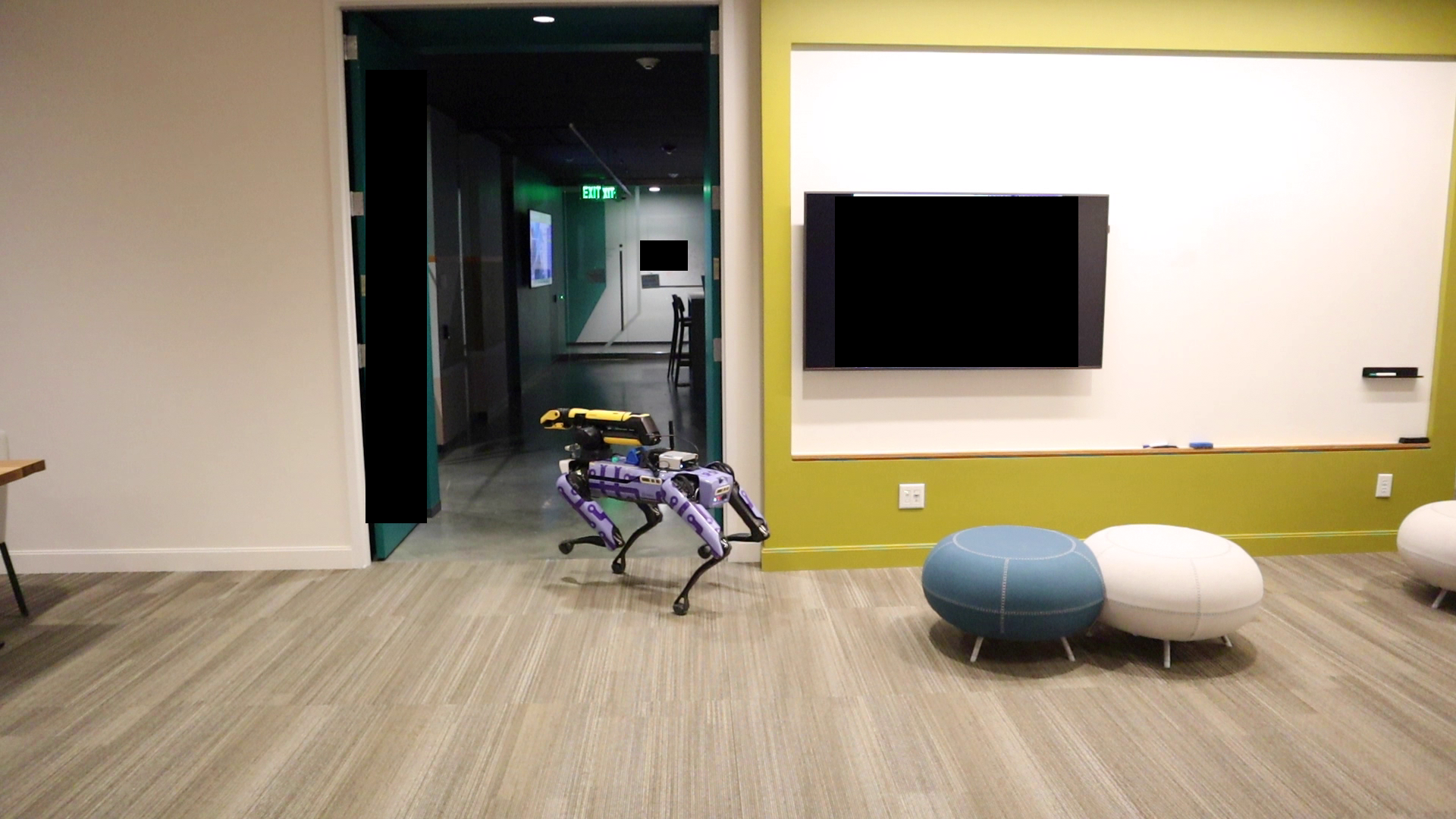}
    \includegraphics[width=0.31\textwidth]{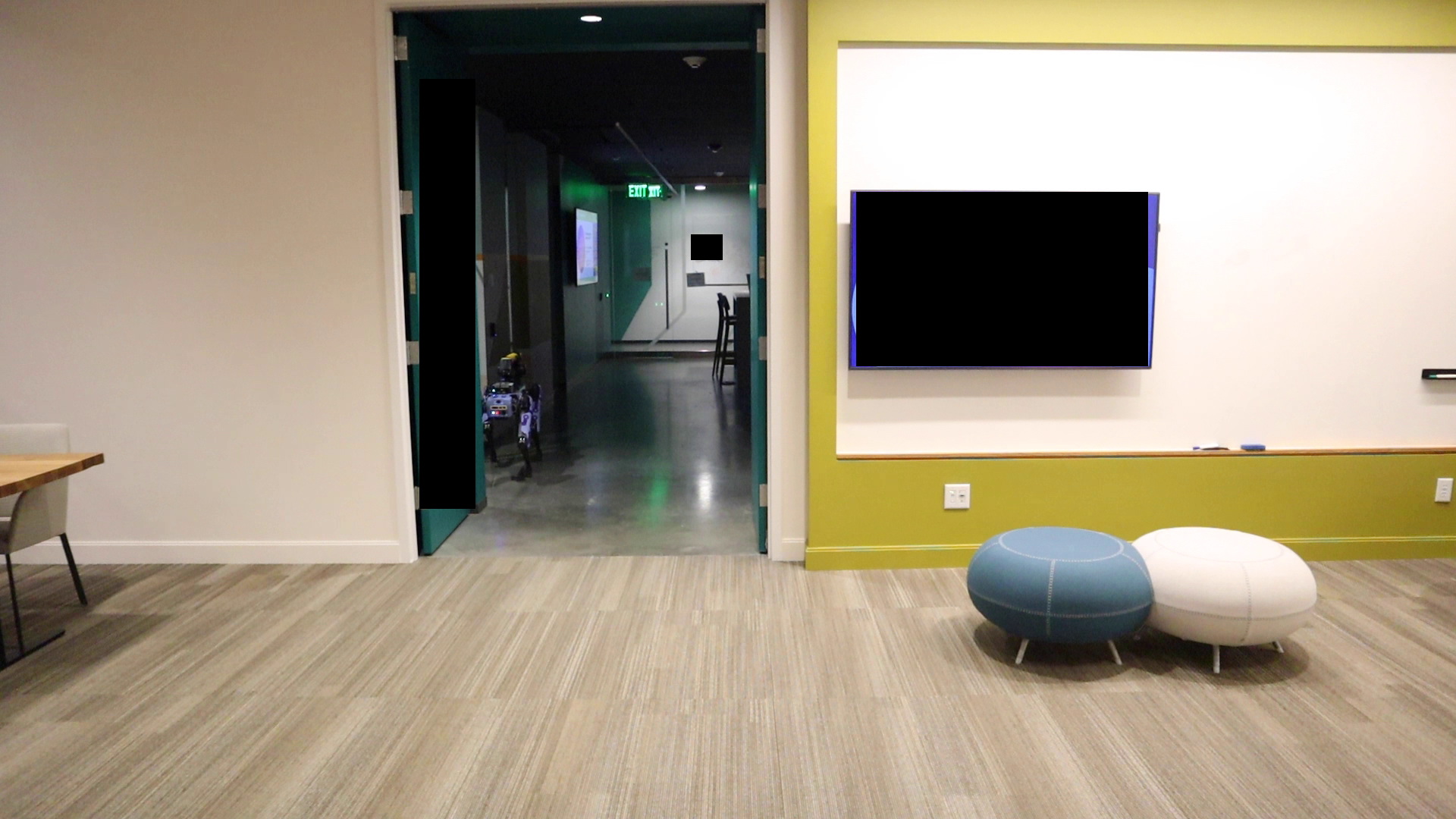}
    \caption{One run of the Spot robot navigating the real-world LAB environment using a kinematically trained policy from AI Habitat. The robot successfully navigates a hallway, moves around furniture and turns into the next hallway before stopping. In contrast, the native BD controllers without a map can only reach visible goals.}
    \label{fig:spot-real}
\end{figure}

\looseness=-1
\label{sec:noise}
\xhdr{How large is the sim2real gap for kinematic and dynamic trained policies?} We evaluate the kinematic and dynamic policies on a Spot robot in the novel LAB environment described in Section \ref{sec:setup}. Note that scans of LAB were not part of training. 
We evaluate 3 seeds of each policy over 5 episodes in the real-world and report the average success rate (SR) and Success weighted by Path Length (SPL) \cite{anderson2018evaluation} in Table \ref{tab:sim2real} (reported as a percentage for readability). Each control type is tested in 15 real-world episodes; %
one run of the Spot robot navigating LAB is shown in Figure \ref{fig:spot-real}. Success in the real-world is measured by computing final distance from the goal position using egomotion estimates provided by the Boston Dynamics SDK.

\begin{table}[h]
	\small
	\vspace{-0.2cm}
	\begin{center}
		\begin{tabularx}{\columnwidth}{ccccccccccc}
			\toprule
			\multicolumn{3}{c}{Train} &  \multicolumn{2}{c}{Simulation} &  \multicolumn{4}{c}{Reality} &  \multicolumn{2}{c}{Sim2Real Gap} \\ 
			\cmidrule(l{4pt}r{4pt}){1-3} \cmidrule(l{4pt}r{4pt}){4-5} \cmidrule(l{4pt}r{4pt}){6-9} \cmidrule(l{4pt}r{4pt}){10-11}
			Simulator & Control & Noise & SR & SPL & SR & SPL & \# Act. & \# Coll. & SR & SPL \\
			\midrule
			Habitat & Dynamic & - & 60.0	& 38.7 & 40.0 & 28.2 & 107.9 & 41.2 & 20.0 & 10.5 \\
			iGibson & Dynamic & - & 20.0 & 14.0 & 67.7 & 46.6 & 76.8 & 12.9 & -47.7 & 32.6 \\
			Habitat & Kinematic & - & 93.3 & 76.9 & \textbf{100.0} & 82.7 & 26.4 & 3.1 & -6.7 & -5.8 \\
			iGibson & Kinematic & - & \textbf{100.0} & \textbf{90.6} & \textbf{100.0} & 83.2 & 33.1 & 4.5 & 0.0 & 7.4 \\
			\midrule
			Habitat & Kinematic & Decoupled & 80.0 & 72.0 & \textbf{100.0} & 87.8 & 27.1 & \textbf{2.5} & -20.0 & -15.8 \\
			Habitat & Kinematic & Coupled & 80.0 & 74.1  & \textbf{100.0} & \textbf{88.8} & \textbf{22.7} & 2.8 & -20.0 & -14.7 \\
			\bottomrule
		\end{tabularx}
	\end{center}
	\caption{Zero-shot sim2real transfer performance for the visual navigation policies. Success rate (SR) and path efficiency (SPL) are high for kinematic policies, while dynamic policies have lower performance due to the dynamics gap between the low-level control in training and the controller on the robot in the real-world.}
	\label{tab:sim2real}
	\vspace{-0.5cm}
\end{table}

\begin{wrapfigure}{r}{0.5\textwidth}
\vspace{-0.3cm}
\begin{tabular}{r r}
   \makecell{\includegraphics[width=0.23\textwidth]{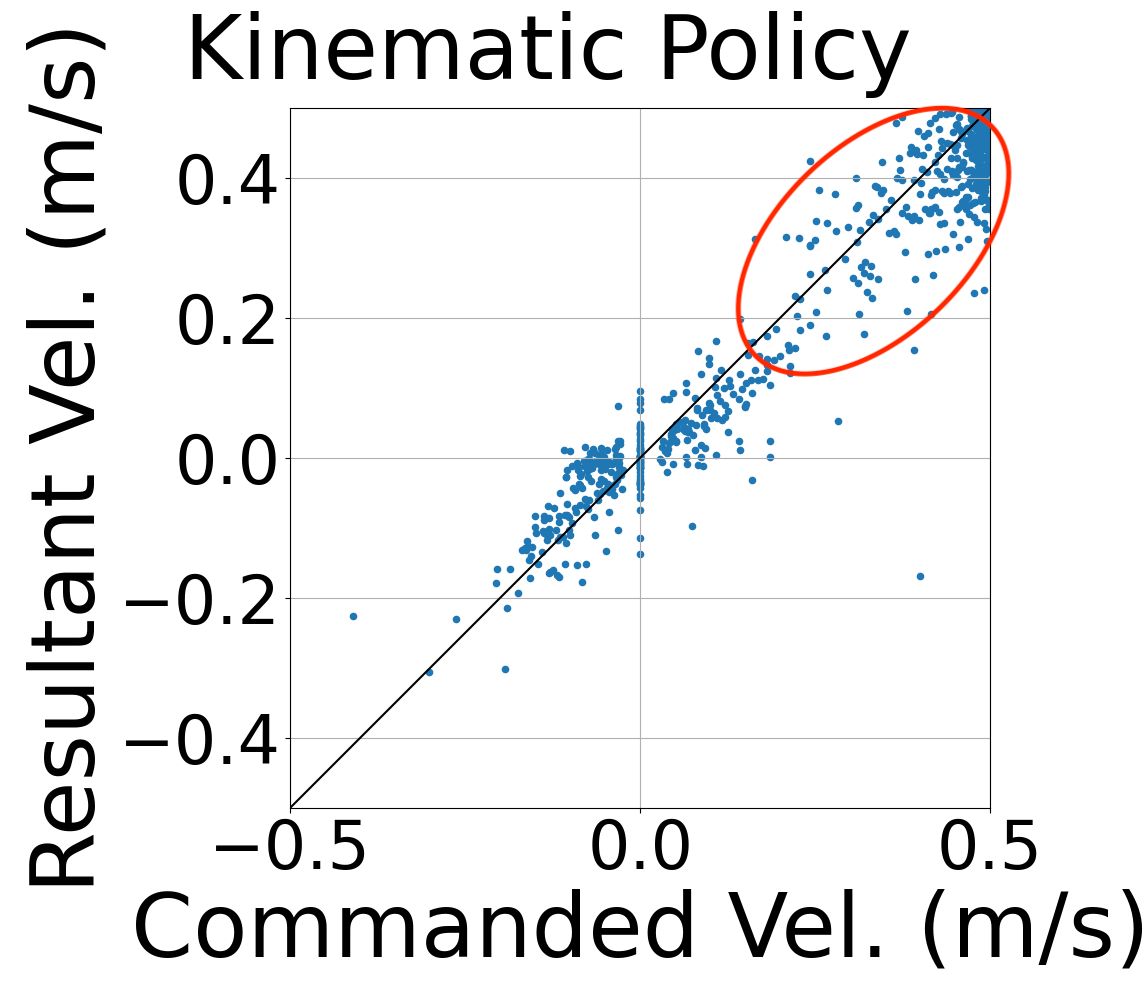}} 
   \makecell{\includegraphics[width=0.23\textwidth]{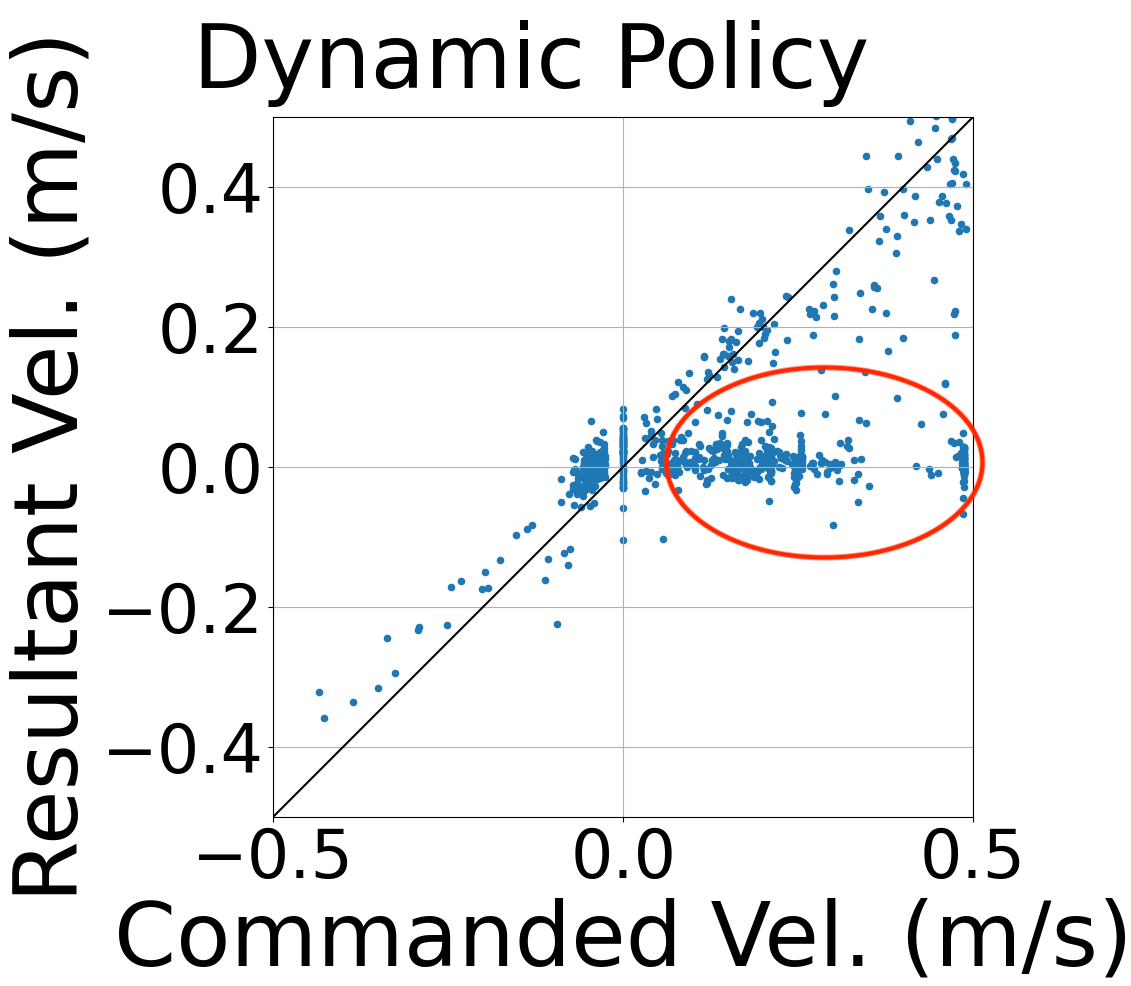}} \\
  \end{tabular}
  \vspace{-0.2cm}
  \caption{Commanded vs. resultant velocities during real-world trajectory rollouts for dynamically and kinematically trained policies.}
  \label{fig:vel_diff}
  \vspace{-0.4cm}
\end{wrapfigure}

As reported in Table \ref{tab:sim2real}, all kinematic policies achieve a high success rate of 100\% and SPL of 82-83\% (rows 3 and 4). On the other hand, the success rate drops to 40-67\% for the dynamic policies (rows 1 and 2). We notice that the dynamic policies typically commanded lower velocities, and often get stuck around obstacles (Figure \ref{fig:vel_diff}). This is shown in the higher number of actions commanded and higher collision count for both dynamic policies; on average, a dynamic policy trained in Habitat took 107.9 actions, and collided 41.2 times (row 1, columns 8 and 9), whereas a kinematic policy also trained in Habitat took 26.4 actions, and collided 3.1 times (row 3, columns 8 and 9). We attribute this to the impoverished experience of the dynamic policies; the policies did not learn robust navigation policies that could avoid obstacles during navigation. Additionally, they overfit to the low-level behavior, which can be unstable at high velocities in sim, but not on hardware. Figure \ref{fig:vel_diff} (left) shows that the kinematic policy commands higher forward velocities, while the dynamic policy commands slower velocities (right), which are often not achieved by the robot likely due to an obstacle. \footnote{Actual velocity is measured using  the Boston Dynamics SDK.} Successfully executed commands appear on the diagonal.

To improve kinematic simulation fidelity, we model actuation noise (difference between commanded and true velocity) on Spot and use it during kinematic training, similar to \cite{truong2021bi}. We collect 6,000 samples of decoupled (linear and angular velocities are actuated separately) and coupled (linear and angular velocities are actuated together) actuation noise. %
The parameters for noise in each dimension, and details about data collection and modeling can be found in the appendix. During training, we sample from the Gaussian distribution for each dimension, and add it to the policy's predicted velocity. 
We see that policies trained with noise also achieve 100\% success in the real-world (rows 5 and 6), and are able to increase path efficiency (SPL) (4.6\% using decoupled (row 4 vs. 5), and 5.6\% using coupled actuation noise (row 4 vs. 6)). The number of collisions and commanded actions are also lower for these policies, compared to kinematic policies trained with no noise (22.7 actions and 2.8 collisions vs. 26.4 actions and 3.1 collisions). %
These improvements are due to the added robustness that training with noise provides -- uncertainty during training forces the policy to take less risky actions resulting in fewer collisions in the real-world.

\vspace{-0.3cm}
\section{Conclusion, Limitations, and Future Work}
\vspace{-0.2cm}
\label{sec:conclusion}
In this work, we study the role of simulation fidelity for sim2real of visual navigation policies on three simulated and one real legged robot. %
Contrary to expectations, we find that higher simulation fidelity does not enable learning better high-level visual navigation policies. Dynamic policies tend to overfit to low-level simulation details, resulting in poor transfer to the real-world. On the other hand, kinematic policies are able to generalize well. These results raise important questions about the need for simulation fidelity for sim2real, especially in abstracted action spaces.

One limitation of this work is that we assume access to a robust `black box' controller on hardware. While most robots come shipped with manufacturer-provided controllers, the level of accuracy may differ between robots, and more robust noise modeling may be needed to better characterize the actuation noise. In the future, we plan to improve the modeling of real-world actuation noise by using a neural network conditioned on previous states and actions of the robot.  Another limitation of this work is that we specifically address only tasks that can be abstracted with high-level actions, like navigation or object rearrangement (pick, place, open, close). We agree that there are many tasks that cannot be learned in a kinematic simulation – e.g. tasks that require reasoning about low-level interactions with the environment, like dexterous manipulation, or low-level walking controllers. It is essential to create high-fidelity simulators for sim2real on such tasks.

\clearpage
\acknowledgments{The Georgia Tech effort was supported in part by NSF, ONR YIP, ARO PECASE. JT was supported by an Apple Scholars in AI/ML PhD Fellowship. The views and conclusions contained herein are those of the authors and should not be interpreted as necessarily representing the official policies or endorsements, either expressed or implied, of the U.S. Government, or any sponsor.

\noindent \textbf{License for dataset used} Gibson Database of Spaces. License at \url{http://svl.stanford.edu/gibson2/assets/GDS_agreement.pdf}
}

\bibliography{bib/main}  %

\begin{thebibliography}{55}
\providecommand{\natexlab}[1]{#1}
\providecommand{\url}[1]{\texttt{#1}}
\expandafter\ifx\csname urlstyle\endcsname\relax
  \providecommand{\doi}[1]{doi: #1}\else
  \providecommand{\doi}{doi: \begingroup \urlstyle{rm}\Url}\fi

\bibitem[Wijmans et~al.(2020)Wijmans, Kadian, Morcos, Lee, Essa, Parikh, Savva,
  and Batra]{ddppo}
E.~Wijmans, A.~Kadian, A.~Morcos, S.~Lee, I.~Essa, D.~Parikh, M.~Savva, and
  D.~Batra.
\newblock {DD-PPO}: Learning near-perfect pointgoal navigators from 2.5 billion
  frames.
\newblock In \emph{International Conference on Learning Representations
  (ICLR)}, 2020.

\bibitem[Savva et~al.(2019)Savva, Kadian, Maksymets, Zhao, Wijmans, Jain,
  Straub, Liu, Koltun, Malik, Parikh, and Batra]{habitat19iccv}
M.~Savva, A.~Kadian, O.~Maksymets, Y.~Zhao, E.~Wijmans, B.~Jain, J.~Straub,
  J.~Liu, V.~Koltun, J.~Malik, D.~Parikh, and D.~Batra.
\newblock Habitat: {A} {P}latform for {E}mbodied {AI} {R}esearch.
\newblock In \emph{International Conference on Computer Vision (ICCV)}, 2019.

\bibitem[Szot et~al.(2021)Szot, Clegg, Undersander, Wijmans, Zhao, Turner,
  Maestre, Mukadam, Chaplot, Maksymets, Gokaslan, Vondrus, Dharur, Meier,
  Galuba, Chang, Kira, Koltun, Malik, Savva, and Batra]{szot2021habitat}
A.~Szot, A.~Clegg, E.~Undersander, E.~Wijmans, Y.~Zhao, J.~Turner, N.~Maestre,
  M.~Mukadam, D.~Chaplot, O.~Maksymets, A.~Gokaslan, V.~Vondrus, S.~Dharur,
  F.~Meier, W.~Galuba, A.~Chang, Z.~Kira, V.~Koltun, J.~Malik, M.~Savva, and
  D.~Batra.
\newblock Habitat 2.0: Training home assistants to rearrange their habitat.
\newblock In \emph{Advances in Neural Information Processing Systems
  (NeurIPS)}, 2021.

\bibitem[Shen et~al.(2021)Shen, Xia, Li, Mart\'in-Mart\'in, Fan, Wang,
  P\'erez-D'Arpino, Buch, Srivastava, Tchapmi, Tchapmi, Vainio, Wong, Fei-Fei,
  and Savarese]{shen2021igibson}
B.~Shen, F.~Xia, C.~Li, R.~Mart\'in-Mart\'in, L.~Fan, G.~Wang,
  C.~P\'erez-D'Arpino, S.~Buch, S.~Srivastava, L.~P. Tchapmi, M.~E. Tchapmi,
  K.~Vainio, J.~Wong, L.~Fei-Fei, and S.~Savarese.
\newblock igibson 1.0: a simulation environment for interactive tasks in large
  realistic scenes.
\newblock In \emph{2021 IEEE/RSJ International Conference on Intelligent Robots
  and Systems (IROS)}, page accepted. IEEE, 2021.

\bibitem[Nvidia(2020)]{isaac}
Nvidia.
\newblock Isaac {S}im.
\newblock \url{https://developer.nvidia.com/isaac-sim}, 2020.

\bibitem[Deitke et~al.(2020)Deitke, Han, Herrasti, Kembhavi, Kolve, Mottaghi,
  Salvador, Schwenk, VanderBilt, Wallingford, et~al.]{deitke2020robothor}
M.~Deitke, W.~Han, A.~Herrasti, A.~Kembhavi, E.~Kolve, R.~Mottaghi,
  J.~Salvador, D.~Schwenk, E.~VanderBilt, M.~Wallingford, et~al.
\newblock Robothor: An open simulation-to-real embodied {AI} platform.
\newblock In \emph{Proceedings of the IEEE/CVF Conference on Computer Vision
  and Pattern Recognition}, pages 3164--3174, 2020.

\bibitem[Gan et~al.(2020)Gan, Schwartz, Alter, Schrimpf, Traer, De~Freitas,
  Kubilius, Bhandwaldar, Haber, Sano, et~al.]{gan2020threedworld}
C.~Gan, J.~Schwartz, S.~Alter, M.~Schrimpf, J.~Traer, J.~De~Freitas,
  J.~Kubilius, A.~Bhandwaldar, N.~Haber, M.~Sano, et~al.
\newblock {ThreeDWorld}: A platform for interactive multi-modal physical
  simulation.
\newblock \emph{arXiv preprint arXiv:2007.04954}, 2020.

\bibitem[James et~al.(2020)James, Ma, Arrojo, and Davison]{james2020rlbench}
S.~James, Z.~Ma, D.~R. Arrojo, and A.~J. Davison.
\newblock Rlbench: The robot learning benchmark \& learning environment.
\newblock \emph{IEEE Robotics and Automation Letters}, 5\penalty0 (2):\penalty0
  3019--3026, 2020.

\bibitem[Xiang et~al.(2020)Xiang, Qin, Mo, Xia, Zhu, Liu, Liu, Jiang, Yuan,
  Wang, Yi, Chang, Guibas, and Su]{xiang2020sapien}
F.~Xiang, Y.~Qin, K.~Mo, Y.~Xia, H.~Zhu, F.~Liu, M.~Liu, H.~Jiang, Y.~Yuan,
  H.~Wang, L.~Yi, A.~X. Chang, L.~J. Guibas, and H.~Su.
\newblock {SAPIEN}: A simulated part-based interactive environment.
\newblock In \emph{The IEEE Conference on Computer Vision and Pattern
  Recognition (CVPR)}, June 2020.

\bibitem[Todorov et~al.(2012)Todorov, Erez, and Tassa]{todorov2012mujoco}
E.~Todorov, T.~Erez, and Y.~Tassa.
\newblock Mujoco: A physics engine for model-based control.
\newblock In \emph{2012 IEEE/RSJ International Conference on Intelligent Robots
  and Systems}, pages 5026--5033. IEEE, 2012.

\bibitem[Freeman et~al.(2021)Freeman, Frey, Raichuk, Girgin, Mordatch, and
  Bachem]{freeman2021brax}
C.~D. Freeman, E.~Frey, A.~Raichuk, S.~Girgin, I.~Mordatch, and O.~Bachem.
\newblock Brax--a differentiable physics engine for large scale rigid body
  simulation.
\newblock \emph{arXiv preprint arXiv:2106.13281}, 2021.

\bibitem[Coumans and Bai(2016)]{coumans2016pybullet}
E.~Coumans and Y.~Bai.
\newblock Pybullet, a python module for physics simulation for games, robotics
  and machine learning.
\newblock 2016.

\bibitem[Ramakrishnan et~al.(2021)Ramakrishnan, Gokaslan, Wijmans, Maksymets,
  Clegg, Turner, Undersander, Galuba, Westbury, Chang,
  et~al.]{ramakrishnan2021habitat}
S.~K. Ramakrishnan, A.~Gokaslan, E.~Wijmans, O.~Maksymets, A.~Clegg, J.~Turner,
  E.~Undersander, W.~Galuba, A.~Westbury, A.~X. Chang, et~al.
\newblock Habitat-matterport 3d dataset (hm3d): 1000 large-scale 3d
  environments for embodied ai.
\newblock \emph{arXiv preprint arXiv:2109.08238}, 2021.

\bibitem[Chang et~al.(2017)Chang, Dai, Funkhouser, Halber, Niessner, Savva,
  Song, Zeng, and Zhang]{chang2017matterport3d}
A.~Chang, A.~Dai, T.~Funkhouser, M.~Halber, M.~Niessner, M.~Savva, S.~Song,
  A.~Zeng, and Y.~Zhang.
\newblock Matterport3d: Learning from rgb-d data in indoor environments.
\newblock \emph{arXiv preprint arXiv:1709.06158}, 2017.

\bibitem[Chang et~al.(2015)Chang, Funkhouser, Guibas, Hanrahan, Huang, Li,
  Savarese, Savva, Song, Su, et~al.]{chang2015shapenet}
A.~X. Chang, T.~Funkhouser, L.~Guibas, P.~Hanrahan, Q.~Huang, Z.~Li,
  S.~Savarese, M.~Savva, S.~Song, H.~Su, et~al.
\newblock {ShapeNet}: An information-rich {3D} model repository.
\newblock \emph{arXiv preprint arXiv:1512.03012}, 2015.

\bibitem[Truong et~al.(2021)Truong, Chernova, and Batra]{truong2021bi}
J.~Truong, S.~Chernova, and D.~Batra.
\newblock Bi-directional domain adaptation for sim2real transfer of embodied
  navigation agents.
\newblock \emph{IEEE Robotics and Automation Letters (RA-L)}, 6\penalty0
  (2):\penalty0 2634--2641, 2021.

\bibitem[Chebotar et~al.(2019)Chebotar, Handa, Makoviychuk, Macklin, Issac,
  Ratliff, and Fox]{chebotar2019closing}
Y.~Chebotar, A.~Handa, V.~Makoviychuk, M.~Macklin, J.~Issac, N.~Ratliff, and
  D.~Fox.
\newblock Closing the sim-to-real loop: Adapting simulation randomization with
  real world experience.
\newblock In \emph{2019 International Conference on Robotics and Automation
  (ICRA)}, pages 8973--8979. IEEE, 2019.

\bibitem[Peng et~al.(2018)Peng, Andrychowicz, Zaremba, and Abbeel]{peng2018sim}
X.~B. Peng, M.~Andrychowicz, W.~Zaremba, and P.~Abbeel.
\newblock Sim-to-real transfer of robotic control with dynamics randomization.
\newblock In \emph{2018 IEEE international conference on robotics and
  automation (ICRA)}, 2018.

\bibitem[Tobin et~al.(2017)Tobin, Fong, Ray, Schneider, Zaremba, and
  Abbeel]{tobin2017domain}
J.~Tobin, R.~Fong, A.~Ray, J.~Schneider, W.~Zaremba, and P.~Abbeel.
\newblock Domain randomization for transferring deep neural networks from
  simulation to the real world.
\newblock In \emph{2017 IEEE/RSJ International Conference on Intelligent Robots
  and Systems (IROS)}. IEEE, 2017.

\bibitem[Kadian et~al.(2020)Kadian, Truong, Gokaslan, Clegg, Wijmans, Lee,
  Savva, Chernova, and Batra]{habitatsim2real20ral}
A.~Kadian, J.~Truong, A.~Gokaslan, A.~Clegg, E.~Wijmans, S.~Lee, M.~Savva,
  S.~Chernova, and D.~Batra.
\newblock Sim2real predictivity: Does evaluation in simulation predict
  real-world performance?
\newblock \emph{IEEE Robotics and Automation Letters (RA-L)}, 2020.

\bibitem[Yokoyama et~al.(2021)Yokoyama, Ha, and Batra]{sct21iros}
N.~Yokoyama, S.~Ha, and D.~Batra.
\newblock Success weighted by completion time: A dynamics-aware evaluation
  criteria for embodied navigation.
\newblock In \emph{2021 IEEE/RSJ International Conference on Intelligent Robots
  and Systems (IROS)}, 2021.

\bibitem[Truong et~al.(2020)Truong, Yarats, Li, Meier, Chernova, Batra, and
  Rai]{truong2020learning}
J.~Truong, D.~Yarats, T.~Li, F.~Meier, S.~Chernova, D.~Batra, and A.~Rai.
\newblock Learning navigation skills for legged robots with learned robot
  embeddings.
\newblock In \emph{International Conference on Intelligent Robots and Systems
  (IROS)}, 2020.

\bibitem[Lee et~al.(2020)Lee, Hwangbo, Wellhausen, Koltun, and
  Hutter]{locomotion_terrain}
J.~Lee, J.~Hwangbo, L.~Wellhausen, V.~Koltun, and M.~Hutter.
\newblock Learning quadrupedal locomotion over challenging terrain.
\newblock \emph{Science Robotics}, 5\penalty0 (47):\penalty0 eabc5986, 2020.
\newblock \doi{10.1126/scirobotics.abc5986}.
\newblock URL
  \url{https://www.science.org/doi/abs/10.1126/scirobotics.abc5986}.

\bibitem[Miki et~al.(2022)Miki, Lee, Hwangbo, Wellhausen, Koltun, and
  Hutter]{locomotion_wild}
T.~Miki, J.~Lee, J.~Hwangbo, L.~Wellhausen, V.~Koltun, and M.~Hutter.
\newblock Learning robust perceptive locomotion for quadrupedal robots in the
  wild.
\newblock \emph{Science Robotics}, 7\penalty0 (62):\penalty0 eabk2822, 2022.
\newblock \doi{10.1126/scirobotics.abk2822}.
\newblock URL
  \url{https://www.science.org/doi/abs/10.1126/scirobotics.abk2822}.

\bibitem[Kumar et~al.(2021)Kumar, Fu, Pathak, and Malik]{kumar2021rma}
A.~Kumar, Z.~Fu, D.~Pathak, and J.~Malik.
\newblock Rma: Rapid motor adaptation for legged robots.
\newblock \emph{Robotics: Science and Systems (RSS)}, 2021.

\bibitem[Garrett et~al.(2020)Garrett, Chitnis, Holladay, Kim, Silver,
  Kaelbling, and Lozano-P{\'e}rez]{garrett2020integrated}
C.~R. Garrett, R.~Chitnis, R.~Holladay, B.~Kim, T.~Silver, L.~P. Kaelbling, and
  T.~Lozano-P{\'e}rez.
\newblock Integrated task and motion planning.
\newblock \emph{arXiv preprint arXiv:2010.01083}, 2020.

\bibitem[Anderson et~al.(2018)Anderson, Chang, Chaplot, Dosovitskiy, Gupta,
  Koltun, Kosecka, Malik, Mottaghi, Savva, et~al.]{anderson2018evaluation}
P.~Anderson, A.~Chang, D.~S. Chaplot, A.~Dosovitskiy, S.~Gupta, V.~Koltun,
  J.~Kosecka, J.~Malik, R.~Mottaghi, M.~Savva, et~al.
\newblock {On Evaluation of Embodied Navigation Agents}.
\newblock \emph{arXiv preprint arXiv:1807.06757}, 2018.

\bibitem[Ramrakhya et~al.(2022)Ramrakhya, Undersander, Batra, and
  Das]{ramrakhya2022habitat}
R.~Ramrakhya, E.~Undersander, D.~Batra, and A.~Das.
\newblock Habitat-web: Learning embodied object-search strategies from human
  demonstrations at scale.
\newblock In \emph{Proceedings of the IEEE/CVF Conference on Computer Vision
  and Pattern Recognition}, pages 5173--5183, 2022.

\bibitem[Chen et~al.(2019)Chen, Gupta, and Gupta]{chen2019learning}
T.~Chen, S.~Gupta, and A.~Gupta.
\newblock Learning exploration policies for navigation.
\newblock \emph{arXiv preprint arXiv:1903.01959}, 2019.

\bibitem[Chaplot et~al.(2020)Chaplot, Gandhi, Gupta, Gupta, and
  Salakhutdinov]{chaplot2020learning}
D.~S. Chaplot, D.~Gandhi, S.~Gupta, A.~Gupta, and R.~Salakhutdinov.
\newblock Learning to explore using active neural slam.
\newblock \emph{arXiv preprint arXiv:2004.05155}, 2020.

\bibitem[Bansal et~al.(2020)Bansal, Tolani, Gupta, Malik, and
  Tomlin]{bansal2020combining}
S.~Bansal, V.~Tolani, S.~Gupta, J.~Malik, and C.~Tomlin.
\newblock Combining optimal control and learning for visual navigation in novel
  environments.
\newblock In \emph{Conference on Robot Learning}, pages 420--429. PMLR, 2020.

\bibitem[Hess et~al.(2016)Hess, Kohler, Rapp, and Andor]{hess2016real}
W.~Hess, D.~Kohler, H.~Rapp, and D.~Andor.
\newblock Real-time loop closure in 2d lidar slam.
\newblock In \emph{ICRA}, 2016.

\bibitem[Nachum et~al.(2019)Nachum, Ahn, Ponte, Gu, and Kumar]{brain}
O.~Nachum, M.~Ahn, H.~Ponte, S.~Gu, and V.~Kumar.
\newblock Multi-agent manipulation via locomotion using hierarchical sim2real.
\newblock \emph{arXiv preprint arXiv:1908.05224}, 2019.

\bibitem[Yu et~al.(2019)Yu, Tan, Bai, Coumans, and Ha]{mso}
W.~Yu, J.~Tan, Y.~Bai, E.~Coumans, and S.~Ha.
\newblock Learning fast adaptation with meta strategy optimization.
\newblock \emph{arXiv preprint arXiv:1909.12995}, 2019.

\bibitem[Li et~al.(2019)Li, Lambert, Calandra, Meier, and Rai]{us}
T.~Li, N.~Lambert, R.~Calandra, F.~Meier, and A.~Rai.
\newblock Learning generalizable locomotion skills with hierarchical
  reinforcement learning.
\newblock \emph{arXiv preprint arXiv:1909.12324}, 2019.

\bibitem[Li et~al.(2021)Li, Calandra, Pathak, Tian, Meier, and
  Rai]{li2021planning}
T.~Li, R.~Calandra, D.~Pathak, Y.~Tian, F.~Meier, and A.~Rai.
\newblock Planning in learned latent action spaces for generalizable legged
  locomotion.
\newblock \emph{IEEE Robotics and Automation Letters}, 6\penalty0 (2):\penalty0
  2682--2689, 2021.

\bibitem[Tan et~al.(2018)Tan, Zhang, Coumans, Iscen, Bai, Hafner, Bohez, and
  Vanhoucke]{tan2018sim}
J.~Tan, T.~Zhang, E.~Coumans, A.~Iscen, Y.~Bai, D.~Hafner, S.~Bohez, and
  V.~Vanhoucke.
\newblock Sim-to-real: Learning agile locomotion for quadruped robots.
\newblock \emph{arXiv preprint arXiv:1804.10332}, 2018.

\bibitem[Peng et~al.(2020)Peng, Coumans, Zhang, Lee, Tan, and
  Levine]{peng2020learning}
X.~B. Peng, E.~Coumans, T.~Zhang, T.-W. Lee, J.~Tan, and S.~Levine.
\newblock Learning agile robotic locomotion skills by imitating animals.
\newblock \emph{Robotics: Science and Systems (RSS)}, 2020.

\bibitem[Rai et~al.(2018)Rai, Antonova, Song, Martin, Geyer, and
  Atkeson]{rai2018bayesian}
A.~Rai, R.~Antonova, S.~Song, W.~Martin, H.~Geyer, and C.~Atkeson.
\newblock Bayesian optimization using domain knowledge on the atrias biped.
\newblock In \emph{2018 IEEE International Conference on Robotics and
  Automation (ICRA)}, pages 1771--1778. IEEE, 2018.

\bibitem[Fu et~al.(2022)Fu, Kumar, Agarwal, Qi, Malik, and
  Pathak]{fu2021coupling}
Z.~Fu, A.~Kumar, A.~Agarwal, H.~Qi, J.~Malik, and D.~Pathak.
\newblock Coupling vision and proprioception for navigation of legged robots.
\newblock \emph{CVPR}, 2022.

\bibitem[Rudin et~al.(2022)Rudin, Hoeller, Reist, and
  Hutter]{rudin2022learning}
N.~Rudin, D.~Hoeller, P.~Reist, and M.~Hutter.
\newblock Learning to walk in minutes using massively parallel deep
  reinforcement learning.
\newblock In \emph{Conference on Robot Learning}, pages 91--100. PMLR, 2022.

\bibitem[Hoeller et~al.(2021)Hoeller, Wellhausen, Farshidian, and
  Hutter]{hoeller2021learning}
D.~Hoeller, L.~Wellhausen, F.~Farshidian, and M.~Hutter.
\newblock Learning a state representation and navigation in cluttered and
  dynamic environments.
\newblock \emph{IEEE Robotics and Automation Letters}, 6\penalty0 (3):\penalty0
  5081--5088, 2021.

\bibitem[Garrett et~al.(2021)Garrett, Chitnis, Holladay, Kim, Silver,
  Kaelbling, and Lozano-P{\'e}rez]{garrett2021integrated}
C.~R. Garrett, R.~Chitnis, R.~Holladay, B.~Kim, T.~Silver, L.~P. Kaelbling, and
  T.~Lozano-P{\'e}rez.
\newblock Integrated task and motion planning.
\newblock \emph{Annual review of control, robotics, and autonomous systems},
  4:\penalty0 265--293, 2021.

\bibitem[Kaelbling and Lozano-P{\'e}rez(2011)]{kaelbling2011hierarchical}
L.~P. Kaelbling and T.~Lozano-P{\'e}rez.
\newblock Hierarchical task and motion planning in the now.
\newblock In \emph{2011 IEEE International Conference on Robotics and
  Automation}, pages 1470--1477. IEEE, 2011.

\bibitem[Lin et~al.(2022)Lin, Wang, Undersander, and Rai]{lin2022efficient}
Y.~Lin, A.~S. Wang, E.~Undersander, and A.~Rai.
\newblock Efficient and interpretable robot manipulation with graph neural
  networks.
\newblock \emph{IEEE Robotics and Automation Letters}, 2022.

\bibitem[Kuindersma et~al.(2016)Kuindersma, Deits, Fallon, Valenzuela, Dai,
  Permenter, Koolen, Marion, and Tedrake]{drc}
S.~Kuindersma, R.~Deits, M.~Fallon, A.~Valenzuela, H.~Dai, F.~Permenter,
  T.~Koolen, P.~Marion, and R.~Tedrake.
\newblock Optimization-based locomotion planning, estimation, and control
  design for the atlas humanoid robot.
\newblock \emph{Autonomous robots}, 40\penalty0 (3):\penalty0 429--455, 2016.

\bibitem[Li et~al.(2019)Li, Geyer, Atkeson, and Rai]{li2019using}
T.~Li, H.~Geyer, C.~G. Atkeson, and A.~Rai.
\newblock Using deep reinforcement learning to learn high-level policies on the
  atrias biped.
\newblock In \emph{ICRA}, pages 263--269. IEEE, 2019.

\bibitem[Zeng et~al.(2020)Zeng, Florence, Tompson, Welker, Chien, Attarian,
  Armstrong, Krasin, Duong, Sindhwani, and Lee]{zeng2020transporter}
A.~Zeng, P.~Florence, J.~Tompson, S.~Welker, J.~Chien, M.~Attarian,
  T.~Armstrong, I.~Krasin, D.~Duong, V.~Sindhwani, and J.~Lee.
\newblock Transporter networks: Rearranging the visual world for robotic
  manipulation.
\newblock \emph{Conference on Robot Learning (CoRL)}, 2020.

\bibitem[Yuan et~al.(2021)Yuan, Paxton, Desingh, and Fox]{yuan2021sornet}
W.~Yuan, C.~Paxton, K.~Desingh, and D.~Fox.
\newblock {SORN}et: Spatial object-centric representations for sequential
  manipulation.
\newblock In \emph{5th Annual Conference on Robot Learning}, 2021.
\newblock URL \url{https://openreview.net/forum?id=mOLu2rODIJF}.

\bibitem[Uni()]{Unitree}
Unitree robotics.
\newblock \url{https://www.unitree.com/}.

\bibitem[spo()]{spot}
Boston dynamics.
\newblock \url{https://www.bostondynamics.com/spot}.

\bibitem[Xia et~al.(2018)Xia, Zamir, He, Sax, Malik, and
  Savarese]{xia2018gibson}
F.~Xia, A.~R. Zamir, Z.~He, A.~Sax, J.~Malik, and S.~Savarese.
\newblock Gibson env: Real-world perception for embodied agents.
\newblock In \emph{CVPR}, 2018.

\bibitem[Yang et~al.(2022)Yang, Zhang, Coumans, Tan, and Boots]{yang2022fast}
Y.~Yang, T.~Zhang, E.~Coumans, J.~Tan, and B.~Boots.
\newblock Fast and efficient locomotion via learned gait transitions.
\newblock In \emph{Conference on Robot Learning}, pages 773--783. PMLR, 2022.

\bibitem[Li et~al.(2021)Li, Won, Ha, and Rai]{li2021model}
T.~Li, J.~Won, S.~Ha, and A.~Rai.
\newblock Model-based motion imitation for agile, diverse and generalizable
  quadupedal locomotion.
\newblock \emph{arXiv preprint arXiv:2109.13362}, 2021.

\bibitem[Hwangbo et~al.(2019)Hwangbo, Lee, Dosovitskiy, Bellicoso, Tsounis,
  Koltun, and Hutter]{hwangbo2019learning}
J.~Hwangbo, J.~Lee, A.~Dosovitskiy, D.~Bellicoso, V.~Tsounis, V.~Koltun, and
  M.~Hutter.
\newblock Learning agile and dynamic motor skills for legged robots.
\newblock \emph{arXiv preprint arXiv:1901.08652}, 2019.

\end{thebibliography}


\begin{thebibliography}{3}
\providecommand{\natexlab}[1]{#1}
\providecommand{\url}[1]{\texttt{#1}}
\expandafter\ifx\csname urlstyle\endcsname\relax
  \providecommand{\doi}[1]{doi: #1}\else
  \providecommand{\doi}{doi: \begingroup \urlstyle{rm}\Url}\fi

\bibitem[Truong et~al.(2020)Truong, Yarats, Li, Meier, Chernova, Batra, and
  Rai]{truong2020learning}
J.~Truong, D.~Yarats, T.~Li, F.~Meier, S.~Chernova, D.~Batra, and A.~Rai.
\newblock Learning navigation skills for legged robots with learned robot
  embeddings.
\newblock In \emph{International Conference on Intelligent Robots and Systems
  (IROS)}, 2020.

\bibitem[Murali et~al.(2019)Murali, Chen, Alwala, Gandhi, Pinto, Gupta, and
  Gupta]{pyrobot2019}
A.~Murali, T.~Chen, K.~V. Alwala, D.~Gandhi, L.~Pinto, S.~Gupta, and A.~Gupta.
\newblock Pyrobot: An open-source robotics framework for research and
  benchmarking.
\newblock \emph{arXiv preprint arXiv:1906.08236}, 2019.

\bibitem[Peng et~al.(2020)Peng, Coumans, Zhang, Lee, Tan, and
  Levine]{peng2020learning}
X.~B. Peng, E.~Coumans, T.~Zhang, T.-W. Lee, J.~Tan, and S.~Levine.
\newblock Learning agile robotic locomotion skills by imitating animals.
\newblock \emph{Robotics: Science and Systems (RSS)}, 2020.

\end{thebibliography}

\end{document}


\maketitle

\section{Robot Details}
\begin{table}[h]\
\vspace{-0.25cm}
\tiny
\centering
\caption{Robot specific parameters used for training and evaluation. The maximum number of steps and velocity limits for each robots are set in proportion to the robot's leg length.}
\label{tab:robot_params}
\resizebox{1.0\columnwidth}{!}{
    \begin{tabular}{l l r c r c r c}
        \toprule
        & Parameter & & A1 & & Aliengo & & Spot \\
		\midrule
        \texttt{1} & Success radius (m)                  & & 0.24 & & 0.32 & & 0.425 \\
		\texttt{2} & Maximum number of steps             & & 326 & & 268 & & 150 \\
		\texttt{3} & Linear velocity limits (m/s)        & $\pm$& 0.23 & $\pm$& 0.28 & $\pm$& 0.50 \\
		\texttt{4} & Angular velocity limits (rad/s)     & $\pm$& 0.14 & $\pm$& 0.17 & $\pm$& 0.3\\
	    \texttt{5} & Leg length (m)                      & & 0.2 & & 0.25 & & 0.44 \\
        \bottomrule
        \vspace{-0.5cm}
    \end{tabular}
    }
\end{table}

\section{Additional Evaluation Results}
\begin{figure}[h]
  \centering%
  \resizebox{\columnwidth}{!}{
  \renewcommand{\tableTitle}[1]{\large{#1}}%
  \setlength{\figwidth}{0.27\columnwidth}%
  \setlength{\tabcolsep}{1.5pt}%
  \renewcommand{\arraystretch}{0.8}%
  \renewcommand{\cellset}{\renewcommand\arraystretch{0.8}%
  \setlength\extrarowheight{0pt}}%

  \hspace{-0.25cm}\begin{tabular}{c c c c}
   \makecell{\includegraphics[width=0.27\textwidth]{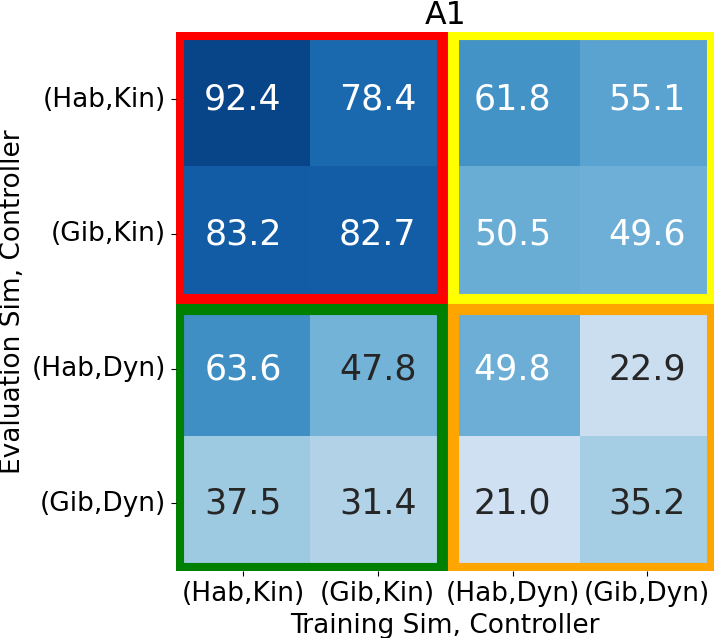}} &
   \makecell{\includegraphics[width=0.27\textwidth]{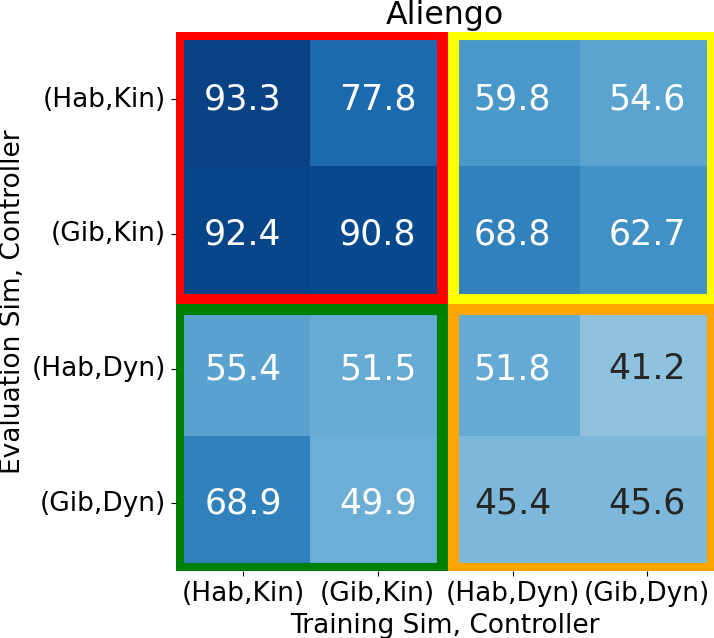}} &
   \makecell{\includegraphics[width=0.27\textwidth]{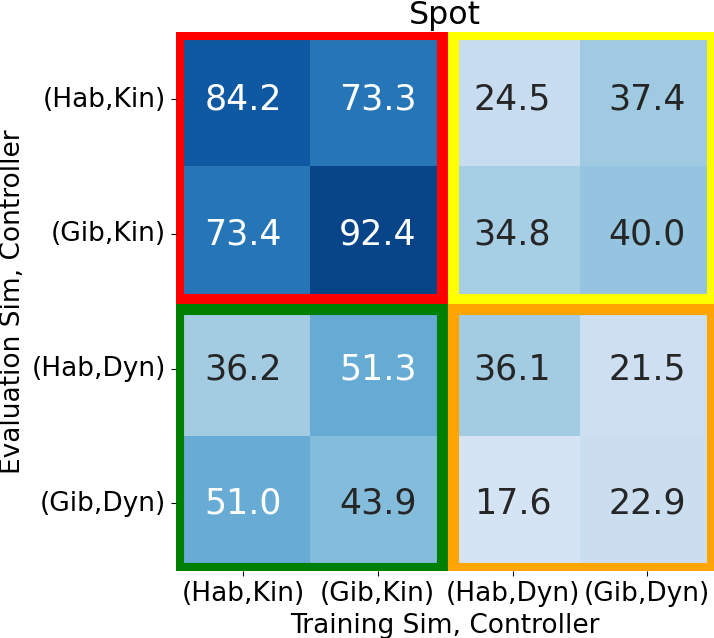}} &
   \makecell{\includegraphics[width=0.042\textwidth]{figures/cbar.png}}
  \end{tabular}}
  \caption{Average success rates for sim2sim and kinematic2dynamic transfer for A1, Aliengo and Spot. Dynamic evaluation in iGibson is performed using the Raibert controller \cite{truong2020learning}. We see that the kinematic trained policies still perform the best overall (\textcolor{red}{red} quadrants), and also often outperforms the dynamic trained policies, even when evaluated using dynamic control (\textcolor{green}{green} quadrants vs. \textcolor{orange}{orange} quadrants). 
} 
  \label{fig:dyn_rai}
  \vspace{-0.3cm}
  \end{figure}

We present additional results using the Raibert controller for evaluation in Figure \ref{fig:dyn_rai} (row 4). The policies are evaluated across 3 seeds, using the  HM3D + Gibson validation split which consists of 1,100 episodes from 110 unique scenes. Our results are consistent with evaluation using the MPC controller-- kinematic trained policies still outperform the dynamic trained policies, \emph{even when evaluated using dynamic control} \footnote{For all robots and training sim/controller except A1, iGibson-Kinematic} (\textcolor{green}{68.9 \% SR} for Aliengo in Habitat, Kinematic vs. \textcolor{orange}{45.4 \% SR} in Habitat, Dynamic, Fig. \ref{fig:dyn_rai}, middle). 

\label{sec:supp}

\section{Actuation Noise Modeling Details}
We collect actuation noise (difference between the commanded and true velocity of the robot) on the Boston Dynamics Spot robot by commanding the robot at a random velocity for 1Hz in an empty room and measuring the final velocity. Noise is collected in a decoupled and coupled manner described below:
\begin{enumerate}
    \item \textbf{Decoupled}: Random velocities ( $\sim \mathcal{U}(-0.5, 0.5) $) are commanded in the forward, lateral, and angular directions \textit{separately}. When collecting data for the forward direction, the sideways direction velocity is commanded zero velocity; the opposite is true when collecting data for the sideways direction. We collect 2,000 datapoints for each direction.
    \item \textbf{Coupled}: Random velocities ( $\sim \mathcal{U}(-0.5, 0.5) $) are commanded in the forward, lateral, and angular directions \textit{at the same time}.
\end{enumerate}

Each dataset contains 6,000 data points, with decoupled data containing 2000 data points for each direction. We choose to model the uncertainty in the robot's actuation with a standard bivariate Gaussian with a diagonal variance similar to \cite{pyrobot2019}. The collected data is used to generate mean and variance parameters for a Gaussian distribution describing the noise in each dimension, as shown in Table \ref{tab:noise_params}. The Gaussian models are then used to inject noise into the the kinematic simulation during training time through the following method: 1) the policy predicts a velocity, 2) the Gaussian distributions for each direction are sampled, 3) the sampled noise is added to the policy's predicted velocity, and 4) the robot's state is updated according to the noisy velocity. 

The two different noise collection approaches aim to study the effects data collection has on the resulting noise model. Our experiments show that both noise models perform better in the real-world than no noise modeling, and coupled noise performs slightly better than decoupled.  

\begin{table}[ht]
\centering
\small
	\begin{center}
    \begin{tabularx}{1.0\columnwidth}{YYYYYYY}
			\toprule
			Noise & $\mu_x$ (m/s) & $\mu_y$ (m/s) & $\mu_\omega$ (deg/s) & $\sigma_x$ (m/s) & $\sigma_y$ (m/s) &  $\sigma_\omega$ (deg/s) \\
			\midrule
			Coupled     & 0.002 &  -0.004 & 0.081 & 0.054 & 0.065 & 2.599 \\
			Decoupled  & 0.002 & -0.001 & -0.029 & 0.036 & 0.044 & 1.468\\
			\bottomrule
		\end{tabularx}
   \end{center}
   \caption{We fit a bivariate Gaussian to the actuation noise collected on a real Spot robot. During kinematic training, we sample from the noise models and inject the realistic actuation noise to the robot's desired final state.}
   \vspace{-0.5cm}
	\label{tab:noise_params}
\end{table}

It is also important to note that while Spot (and other legged robots) can move in all directions, these robots are not necessarily omnidirectional platforms, since they cannot move in all directions \textit{equally well}. 
\begin{figure}[h]
    \centering
    \includegraphics[width=0.7\columnwidth]{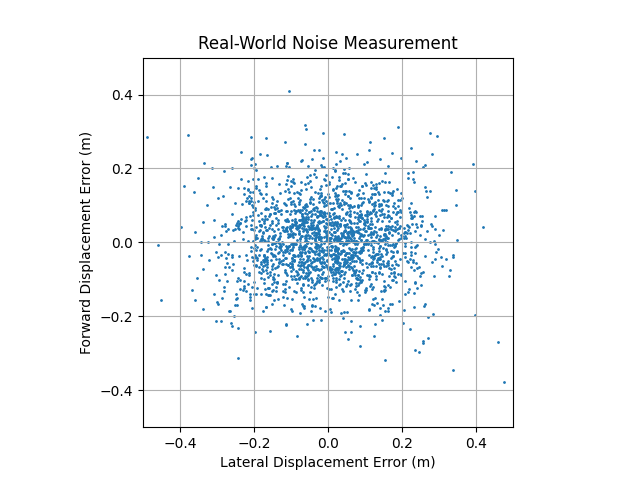}
    \caption{\small We collect displacement errors on Spot in forward and lateral directions. The standard deviation for the displacement errors between the forward and lateral directions are large and asymmetric, demonstrating that the robot is not perfectly omnidirectional.}
    \label{fig:disp_error}
\end{figure}
To illustrate this, we collect displacement errors on Spot in forward and lateral directions while commanding random desired CoM movements (Figure \ref{fig:disp_error}). If the robot were perfectly omnidirectional, we would expect the means and variances for the forward and lateral direction to be small and the same. While the mean error in both directions is close to 0, the standard deviations in the forward and lateral directions are significantly larger and asymmetric. In the forward direction, the standard deviation is 0.097 meters, and in the lateral direction it is 0.139 meters. This tracking error, which increases with commanded velocity, motivated the choice of saturating commanded desired velocity at 0.5 m/s. 
This behavior is observed on the Spot robot using Boston Dynamics walking controllers, which is a very good, highly tuned controller for the robot. We would expect any open-sourced controller which is not tuned for a particular robot to only be worse.
\section{Additional Low-level Controller Details}
We use two different kinds of low-level controllers in our work-- an expert-designed Raibert controller from \cite{truong2020learning} (modified to allow for lateral movement), and a model-predictive control (MPC) controller from \cite{peng2020learning}. The Raibert controller takes in desired CoM velocities $(v_{x\_des}, v_{y\_des}, \omega_{des})$ from the high-level policy to calculate the desired foot placement location, following equations 1-3 from \cite{truong2020learning}. The footstep trajectory is followed using inverse kinematics. The MPC controller uses a contact schedule to determine each leg's contact state and compute the optimal joint torque for each leg.
\section{Additional Policy Details}
\subsection{High-level policy parameters.}
We use PPO with Generalized Advantage Estimation (GAE). We use a discount factor of 0.99, and GAE parameter of 0.95. We use the Adam optimizer, with a learning rate of 2.5e-4. We run 8 agents in parallel (in different environments) per GPU, and each agent collects a rollout of 128 frames of experience. We use 8 GPUs, for a total of 64 parallel workers.
\subsection{Reward function.}
Our reward function is derived from \cite{truong2020learning}, with
an added penalty for backward velocities, which can lead to collisions and hurts performance. Specifically, our reward function is defined as:
\begin{align}
    r_t(a_t,s_t) = R_{geo} + R_{coll} + R_{fall} + R_{success} + R_{slack} + R_{backward}
\end{align}
$R_{geo}$ is a shaped reward, denoting the change in geodesic distance to the goal between two timesteps. \\
$R_{coll}$ is a penalty for collisions. We set the collision penalty to -0.03. \\
$R_{fall}$ is a penalty if the robot falls over. We set the falling penalty to -5.0, and terminate the episode.\\
$R_{success}$ is the terminal reward for completing the episode. We set the terminal reward to 10.0.\\
$R_{slack}$ is a slack penalty used to encourage the robot to reach the goal as fast as possible. We set the slack penalty to -0.002.\\
$R_{backward}$ is a penalty for moving backwards, as moving backwards can lead to collisions. We set the backwards penalty to -0.03.

\subsection{Dynamic Simulation Overfitting Details.}
We define overfitting as the drop in performance when testing on a different controller and/or simulator than training. This is a natural generalization of the standard definition of overfitting in supervised learning (accuracy on IID training dataset - accuracy on IID testing dataset). We train dynamic policies for all three robots to congergence (Figure \ref{fig:train_plots})\\
In Table \ref{tab:overfit}, we show the success rate on Habitat-Dynamic (training scenario) - success rate on iGibson-Dynamic (testing scenario) for all 3 robots. Note that these are all evaluations on the same houses/scenes/environments (from a held-out evaluation set) and the only factor changing is the simulator. We can clearly see that the gap is always positive, indicating that policies trained on Habitat-Dynamic perform worse when evaluated on iGibson-Dynamic compared to evaluation on Habitat-Dynamic. As can be expected, in all but one case, the performance gaps are increasing with more RL training, though this is not strictly necessary. A well-trained high-level policy can learn to reason intelligently about navigation (even with dynamic controllers), and then perform well across simulators.

\begin{table}[h]
\centering
\small
	\begin{center}
    \begin{tabularx}{0.5\columnwidth}{YYY}
			\toprule
			Robot & Steps of experience & Performance gap (\%)\\
			\midrule
			\multirow{3}{*}{A1}       & 12M &  7.8 \\
		                              & 25M &  26.7 \\
		                              & 50M &  27.2 \\
		    \midrule
		    \multirow{3}{*}{AlienGo}  & 12M &  5.7 \\
		                              & 25M &  5.4 \\
		                              & 50M &  8.1 \\
		    \midrule
		    \multirow{3}{*}{Spot}     & 12M &  14.0 \\
		                              & 25M &  21.6 \\
		                              & 50M &  15.6 \\
			\bottomrule
		\end{tabularx}
   \end{center}
   \caption{We measure the performance gap for dynamic policies between evaluations in Habiat-Dynamic and iGibson-Dynamic. The gap is always postive, and in all cases but one, the performance gaps increase with more RL training, demonstrating that the dynamic policies overfit to the simulator and controller it was trained on.}
   \vspace{-0.5cm}
	\label{tab:overfit}
\end{table}

\newpage

\bibliography{bib/main}  %